\documentclass[11pt]{article}

\usepackage{graphicx} 
\usepackage{amsmath, amssymb,amsthm}
\usepackage{algorithm}
\usepackage{subcaption}    %
\usepackage{float}
\usepackage{algorithmic}
\usepackage[numbers,sort&compress]{natbib} 
\usepackage{authblk} 
\usepackage[margin=1in]{geometry} 
\usepackage{hyperref}
\usepackage{url}
\usepackage{booktabs}
\usepackage{multirow}
\usepackage{xcolor}
\usepackage{pifont}

\newcommand{\cmark}{\textcolor{green!60!black}{\ding{51}}} 
\newcommand{\xmark}{\textcolor{red!80!black}{\ding{55}}}
\newtheorem{definition}{Definition}
\newtheorem{theorem}{Theorem}
\newtheorem{lemma}{Lemma}

\usepackage{mathtools}
\usepackage{bm}

\title{Toward Temporal Causal Representation Learning with Tensor Decomposition}

\author[1]{Jianhong Chen}
\author[2]{Meng Zhao}
\author[3]{Mostafa Reisi Gahrooei}
\author[1]{Xubo Yue\thanks{Corresponding Author: \texttt{x.yue@northeastern.edu}}}

\affil[1]{Department of Mechanical \& Industrial Engineering, Northeastern University, Boston, MA, USA}
\affil[2]{Department of Industrial and Systems Engineering, Lehigh University, Bethlehem, PA, USA}
\affil[3]{Department of Industrial and Systems Engineering, University of Florida, Gainesville, FL, USA}

\begin{document}

\maketitle
\begin{abstract}

Temporal causal representation learning has been a powerful tool for uncovering complex patterns in observational studies, which are often represented as low-dimensional time series. However, in many real-world applications, data are high-dimensional with varying input lengths, and naturally take the form of irregular tensors. To analyze such data, irregular tensor decomposition is critical for extracting meaningful clusters that capture essential information. In this paper, we focus on modeling causal representation learning based on the transformed information. First, we present a novel causal formulation for a set of latent clusters. We then propose \texttt{CaRTeD}, a joint‐learning framework that integrates temporal causal representation learning with irregular tensor decomposition. Notably, our framework provides a blueprint for downstream tasks using the learned tensor factors, such as modeling latent structures and extracting causal information, and offers a more flexible regularization design to enhance tensor decomposition. Theoretically, we show that our algorithm converges to a stationary point. More importantly, our results fill the gap in theoretical guarantees for the convergence of state‐of‐the‐art irregular tensor decomposition. Experimental results on synthetic and real-world electronic health record (EHR) datasets (MIMIC-III) with extensive benchmarks from both phenotyping and network recovery perspectives demonstrate that our proposed method outperforms state-of-the-art techniques and enhances the explainability of causal representations.
\end{abstract}

\section{Introduction}

Causal Representation Learning (CRL), also known as causal discovery (CD), aims to infer the underlying causal structure among a set of variables. The causal structure is often represented as a Directed Acyclic Graph (DAG), which explicitly avoids circular dependencies between causes and effects. CRL has been applied across diverse domains, such as reconstructing gene regulatory networks from high-throughput data \citep{Gao2022CausalRecSys} and elucidating molecular pathways in genomic medicine \citep{Kelly2022Review}. One particular application is the construction of \textit{causal phenotype networks} \citep{rosa2011inferring}, which use quantitative methods to infer the underlying phenotypic relationships to predict the effects of interventions. For example, in clinical practice, for a patient with heart failure and comorbid kidney disease, extensive examination may reveal a causal relationship in which kidney disease can lead to heart failure. Therefore, by modeling the causal relationship, we can assess how an intervention targeting kidney disease may influence the progression or severity of heart failure. 

The widespread adoption of electronic health record (EHR) systems has generated substantial volumes of clinical data, providing a valuable resource for a broad range of research studies. In the healthcare system, strategically leveraging and analyzing EHR data can enhance operational efficiency and enable more cost-effective treatment and management plans. In recent years, a key use of EHR data is computational phenotyping, the goal of which is to derive more nuanced, data-driven characterizations of disease \citep{che2015DCP}. Computational phenotyping seeks to identify meaningful clusters or patterns in patient data, such as diagnosis codes, to define clinical conditions. Unsupervised low-rank techniques, such as tensor factorization, have shown considerable promise by representing complex patient data as third-order tensors \citep{HO2014PG,Wang2015Rubik}. For example, we can model the EHR dataset as a tensor with three modes: patients, diagnoses, and visits. The tensor representation of EHR data not only encodes patient trajectories over time but also highlights the unparalleled depth of clinical information. Moreover, many modern data sources are inherently high-dimensional, making tensors their most natural representation. Developing causal representations within a tensor analysis framework therefore constitutes an important new research direction. \textit{However, existing causal structure learning methods are typically designed for flat observational study.} The key problem is to extend causal-structure learning to tensor data and to integrate tensor-specific techniques, overcoming their current limitation to low-dimensional inputs and enabling their use on inherently high-dimensional datasets.

\textit{Methods for learning meaningful clusters or patterns and the causal structure among them from tensor data are therefore essential}. In this work, we integrate causal structure learning with tensor decomposition based data mining tools. As a concrete example, we construct causal phenotype networks from EHR data. Tensor factorization based methods are typically divided into static and temporal approaches \citep{Becker2023}. In static phenotyping, all visits for each patient are collapsed into a regular third‐order tensor, often defined over patient, diagnosis, and medication modes, and then analyzed via CANDECOMP/PARAFAC (CP) decomposition to uncover co‐occurrence patterns \citep{HO2014PG, Wang2015Rubik, Kim2017Discriminative}. By contrast, temporal phenotyping preserves the longitudinal sequence of clinical events, modeling each patient’s record as a temporally irregular tensor to extract phenotypes and their dynamic trajectories over time. For instance, an EHR dataset may include $K$ patients, each characterized by $J$ clinical variables measured over $I_k$ encounters for the $k^{th}$ patient, where the number of visits $I_k$ varies across individuals. In this situation, CP decomposition no longer applies. To handle such irregular tensors, a more flexible model known as PARAFAC2 factorization \citep{Hars1972b} has been applied for temporal phenotyping, where each phenotype represents a set of co-occuring clinical features (e.g., diagnoses). In this paper, we use temporal phenotyping to infer the underlying temporal causal structure among those phenotypes. However, simply applying temporal phenotyping decomposition and a causal discovery separately is not feasible, as the decomposition results may lack accuracy without causal-informed regularization, and the quality of causal structure learning is also influenced by the outcomes of the tensor decomposition. Hence, it is necessary to jointly learn both temporal phenotyping and causal structure. Accordingly, we propose a unified framework that integrates these two tasks in a principled manner.

\begin{figure}
    \centering
    \includegraphics[width=0.8\linewidth]{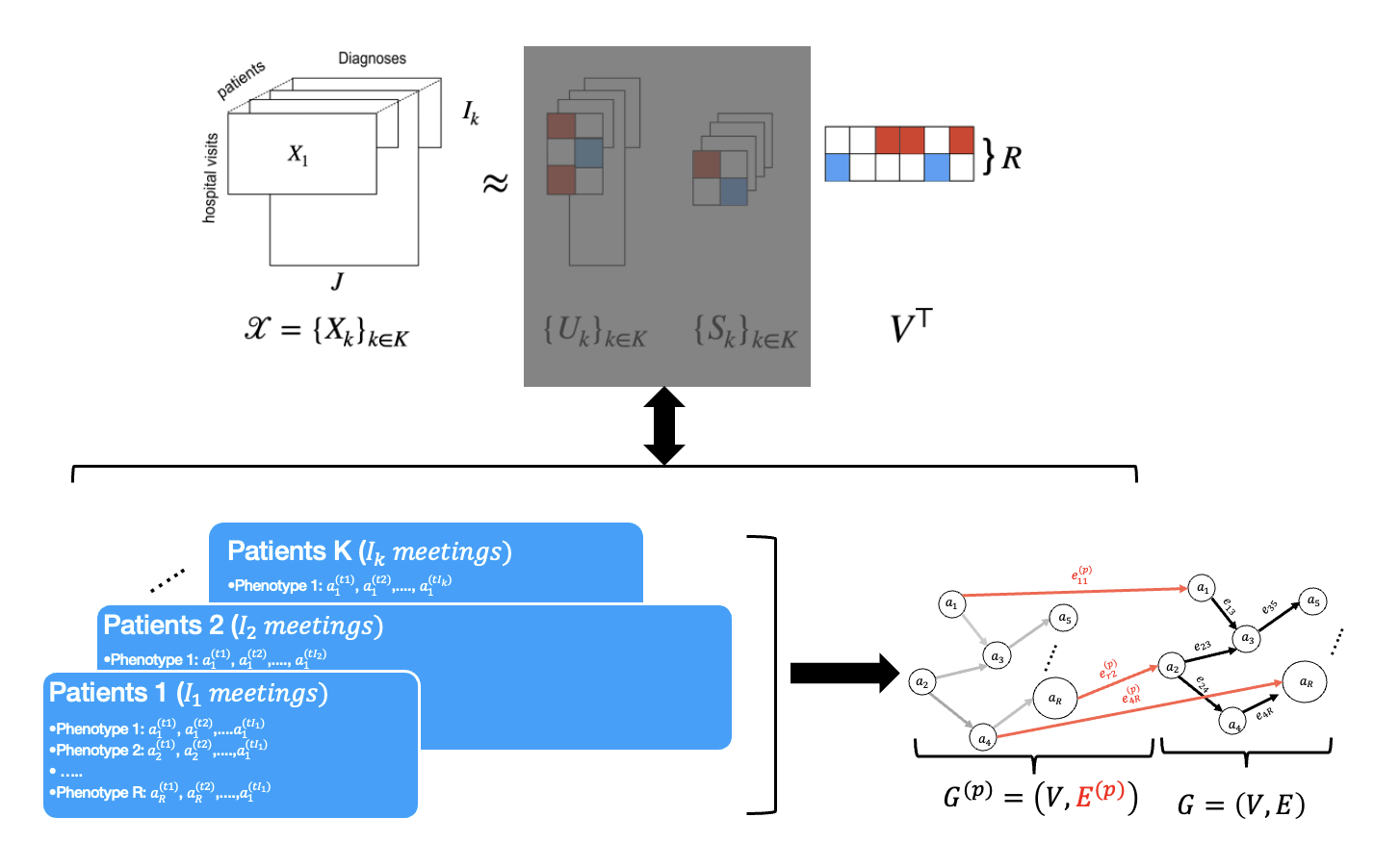}
    \caption{Overview of causal relationships in the PARAFAC2 decomposition. The graph with red edges, $E^{(p)}$, represents the temporal network that captures lag-$p$ effects, whereas the graph with black edges, $E$, represents the contemporaneous network. Formal definitions of these graphs and the PARAFAC2 decomposition are provided in Section \ref{sec:pf}.}
    \label{fig:CaRTed_causal}
\end{figure}

\paragraph{Research Gaps and Our Contributions}  
In this paper, we bridge causal-structure learning with irregular tensor decomposition to learn a temporal causal phenotype network from EHR data. To highlight our contributions, and in contrast to existing irregular tensor decomposition methods, our approach can tackles two intertwined causal questions for each discovered phenotype cluster: (i) the contemporaneous network, asking whether phenotype $i$ has an immediate, direct causal influence on phenotype $j$ among the $R$ phenotypes, and (ii) the temporal network, asking whether phenotype $i$ observed at an earlier time $t-\tau$ ($\tau>0$) causally affects phenotype $j$ at time $t$. Fig.~\ref{fig:CaRTed_causal} illustrates these causal relationships in the context of PARAFAC2 decomposition. To tackle these problems, we must overcome challenges from two complementary perspectives: one arising from causal‐structure learning over latent variables in high-dimensional data with irregular time steps, and the other from extending tensor-decomposition frameworks beyond mere reconstruction to support downstream causal analysis. Specifically, from the causal‐structure‐learning perspective, current methods cannot directly extract meaningful information from irregular tensor data. From the tensor‐decomposition perspective, existing approaches focus solely on decomposition quality and do not support downstream tasks, such as the structure modeling and causal analysis. Moreover, these methods do not incorporate meaningful causal information into the tensor‐decomposition learning process. Our contributions can be summarized as follows:

\begin{itemize}
    \item We propose a novel joint learning framework that unifies temporal causal phenotype network inference and computational phenotyping. Technically, we tackle key challenges within the tensor‐decomposition framework, laying the groundwork for future research on related tasks:
    \begin{itemize}
      \item Existing constraints are insufficient, as they regulate only a single factor. We instead propose a combined constraint to better enforce joint structure.
      \item Most irregular tensor‐decomposition methods assume that constraints are known. Our framework can be used to handle latent or dynamic constraints directly.
    \end{itemize}
    \item We provide a theoretical convergence analysis for the resulting non-convex optimization problem with non-convex constraints.  
    \item Through extensive simulations on diverse benchmarks and evaluation metrics, we demonstrate that our method is scalable and accurately recovers both the underlying phenotypes and their causal relationships.  
    \item We apply our methodology to the MIMIC-III dataset to extract phenotypes and infer a causal phenotype network, demonstrating that our joint learning framework achieves superior accuracy compared to the benchmark, the two-step learning approach.
    \item To the best of our knowledge, this is the first study examining temporal causal phenotype networks within an irregular tensor decomposition framework. Our code is publicly available on GitHub\footnote{\url{https://github.com/PeChen123/CaRTed}}.

\end{itemize}

\paragraph{Literature Review:} 

In this section, we review related work in three primary areas: unsupervised low-rank approximation methods for computational phenotyping, causal discovery techniques, and causal phenotype networks.

Tensor factorization techniques are effective for extracting phenotypes because EHR data can be represented as matrices or higher‐order tensors. For \emph{static phenotyping}, EHR records are aggregated over time (i.e., all visits for each patient are combined) and organized into a regular third‐order tensor, which is then analyzed using CANDECOMP/PARAFAC (CP) decomposition \citep{HO2014PG}. For example, \citet{Wang2015Rubik} builds a tensor with patient, diagnosis, and medication modes. Each patient is represented by a matrix of cumulative diagnosis and prescription counts across all visits, and factorizes it via CP. Similarly, \citet{Kim2017Discriminative} arranges EHR data as a diagnosis–prescription co‐occurrence tensor and apply CP decomposition. These approaches assume a \emph{regular} tensor, where each mode’s dimensions align across patients, and thus break down on \emph{irregular} tensors arising from varying numbers of visits or measurement frequencies. To address irregularity, several PARAFAC2‐based methods have been proposed. Prior to PARAFAC2, \citet{CHi2021heter} applied dynamic time warping to align irregular time modes before CP decomposition. PARAFAC2 itself accommodates one mode with varying dimensions, and has been extended for EHR phenotyping in multiple works: \citet{Perros2017SPARTan} introduced SPARTan, a scalable PARAFAC2 algorithm for large, sparse temporal EHR tensors; \citet{Afshar2018COPA} enhanced SPARTan with temporal smoothness, nonnegativity, and sparsity constraints (COPA); \citet{Ren2020Tensor} imposed low‐rankness constraints to improve robustness to missing or noisy entries (REPAIR); and \citet{Yin2020LogPar} developed LogPar, a logistic PARAFAC2 model for binary, irregular tensors with missing data. More recently, \citet{Ren2022MULTIPAR} embedded PARAFAC2 within a supervised multi‐task learning framework (MULTIPAR), further enhancing its applicability to heterogeneous EHR datasets.

Methods for learning causal structure fall into three main categories: constraint-based, score-based, and hybrid approaches. As shown by \citet{scutari2019learnsbetter}, score-based methods often achieve higher accuracy without extra computational cost compared to either constraint-based or hybrid-based approaches. Score-based methods consist of two key steps: model scoring and model search. These methods cast the search for a causal graph \(G\) as an optimization problem over a scoring function \(S\). Specifically, \citet{Peter2017} define $\hat{G} = \arg\min_{G} S(D, G),$ where \(D\) denotes the empirical data for variables \(\boldsymbol{x}\). A canonical score-based framework is the Bayesian network (BN), which models contemporaneous causal relationships but may overlook temporal dynamics. To capture time-lagged effects, Dynamic Bayesian Networks (DBNs) were introduced by \citet{Murphy2002}. However, structure learning is a combinatorial optimization problem and finding a globally optimal network is NP‐hard. To address this, \citet{Zheng2018} reformulated acyclicity as a differentiable algebraic constraint and embedded it in a continuous optimization problem, an approach later extended by \citet{ng2019gau}, \citet{lachapelle2020gb}, and \citet{Petkov_2022}. More recently, \citet{pamfil2020dynotears} generalized this continuous optimization framework to temporal causal discovery.

As an extension of causal discovery, causal phenotype networks (CPNs) aim to infer directional causal relationships among phenotypic traits derived from clinical or genetic data. \citet{hidalgo2009dynamic} first introduced phenotypic disease networks (PDNs) to map comorbidity correlations across millions of medical records, yielding undirected associations among diseases. \citet{rosa2011inferring} advanced this approach by integrating structural equation models (SEMs) with quantitative trait loci (QTL) information to disentangle direct and indirect causal effects between phenotypes. Building on these foundations, \citet{chaibub2010causal} proposed causal graphical models that jointly infer phenotype–phenotype networks and their underlying genetic architectures using conditional Gaussian regression frameworks. More recently, \citet{shen2021novel} tailored causal discovery to EHR data via novel transformations and bootstrap aggregation, enhancing the stability and clinical consistency of recovered directed acyclic graphs in chronic disease cohorts. However, all these approaches does not account for the tensor structure of the data. Thus, learning temporal causal phenotype networks from EHR data is of critical importance.

Despite advances in tensor-based computational phenotyping and score-based causal representation learning, a clear gap remains between these paradigms. Consequently, we have summarized the differences between our method and the most relevant tasks described above in Table~\ref{tab:method_com}. \textbf{To our knowledge, no prior work has integrated causal discovery into tensor decomposition frameworks for causal phenotype networks}. Our proposed framework fills this gap by embedding causal-structure learning directly within the tensor factorization process, enabling the handling of unknown structural constraints in irregular tensor data.

\begin{table}[h]
  \centering
  \caption{Comparison between the most relevant methods and our proposed method}
  \label{tab:method_com}
  \begin{tabular}{lcccccc}
    \toprule
   & \texttt{QTLnet}\citep{chaibub2010causal} &\texttt{DYN}\citep{pamfil2020dynotears} & \texttt{C-SEM}\citep{rosa2011inferring}& \texttt{COPA}\citep{Afshar2018COPA} & \texttt{CD-EHR}\citep{shen2021novel}
   & \texttt{CaRTeD} \\
    \midrule
     Theoretical Analysis        & \cmark       & \xmark & \xmark & \xmark  & \xmark & \cmark         \\
    Static Causal Structure      & \cmark & \cmark &  \cmark  & \xmark   & \cmark & \cmark               \\
    Temporal Causal Structure & \xmark & \cmark &  \xmark  & \xmark   & \xmark & \cmark               \\
     Computational Phenotype & \xmark    & \xmark  &  \xmark  &\cmark & \xmark & \cmark               \\
     Handle Irregular Tensors & \xmark   &  \xmark  &  \xmark &   \cmark & \cmark & \cmark \\
   \bottomrule
  \end{tabular}
\end{table}

\section{Problem Formulation}
\label{sec:pf}

In this section, we first introduce the concepts of tensor operations and irregular tensors. We then describe the classical PARAFAC2 factorization, the constrained PARAFAC2 (COPA) and its practical application to temporal EHR-based phenotyping. Next, we present the formulation of dynamic Bayesian networks and graph notations. Finally, we describe the problem formulation of our proposed Causal Representation learning with irregular Tensor Decomposition (CaRTeD) framework for learning the causal phenotype network. 

\subsection{Tensor Operations and Irregular Tensors}
In this article, the higher‐order tensors are denoted by calligraphic letters \(\mathcal{X}\). Scalars, vectors, and matrices are denoted by lowercase or capital letters (e.g.,\ \(x\) or \(X\)). \textit{Slices} refer to two-dimensional sections of a tensor, defined by fixing all modes but two indices. There are horizontal, lateral, and frontal slices of a third-order tensor \(\mathcal{X}\). For example, the frontal slices are defined by $ \mathcal{X}(:,:,k)$ $(k = 1,2,\dots,K)$, which are simply denoted by \({X}_{k}\). A mode-k \textit{fiber} of a tensor is a subarray of a tensor that is obtained by fixing all the mode indices but mode $k$. Tensor matricization along a mode (say mode $k$) converts a tensor into a matrix whose columns are the mode-k fibers of the tensor and is typically denoted by \({X}_{(k)}\). The symbols \(\odot\), \(\otimes\), and \(*\) denote the Khatri-Rao, Kronecker product, and Hadamard products of two matrices, respectively. The Frobenius norm of a tensor \(\mathcal{X}\) equals the Frobenius norm of any unfolded format of \(\mathcal{X}\), denoted as $\|\mathcal{X}\|_F = \|{X}_{(n)}\|_F$ $(n = 1, \dots, N)$. The \(\ell_1\) norm of a tensor \(\mathcal{X}\) is denoted as \(\|\mathcal{X}\|_1\), calculated as the sum of the absolute values of its entries.

An \textit{irregular tensor} refers to a multidimensional data structure where the dimensions vary across at least one of its modes. For example, the EHR data can be represented as $\mathcal{X} = \{{X}_k \in \mathbb{R}^{I_k \times J}\}_{k=1}^K$, a set of $K$ matrices each encoding one patient’s information. Each matrix comprises $J$ clinical features (e.g., diagnoses) collected over $I_k$ visits. The Frobenius norm and $\ell_1$ norm of an irregular tensor are defined as the sum of the corresponding norms of its constituent frontal slices, respectively:
\[
  \|\mathcal{X}\|_F = \sum_{k=1}^K \|{X}_k\|_F,
  \qquad
  \|\mathcal{X}\|_1 = \sum_{k=1}^K \|{X}_k\|_1.
\]

\subsection{PARAFAC2 Factorization and Temporal EHR Phenotyping}
\begin{figure}[H]
    \centering
    \includegraphics[width=0.6\linewidth]{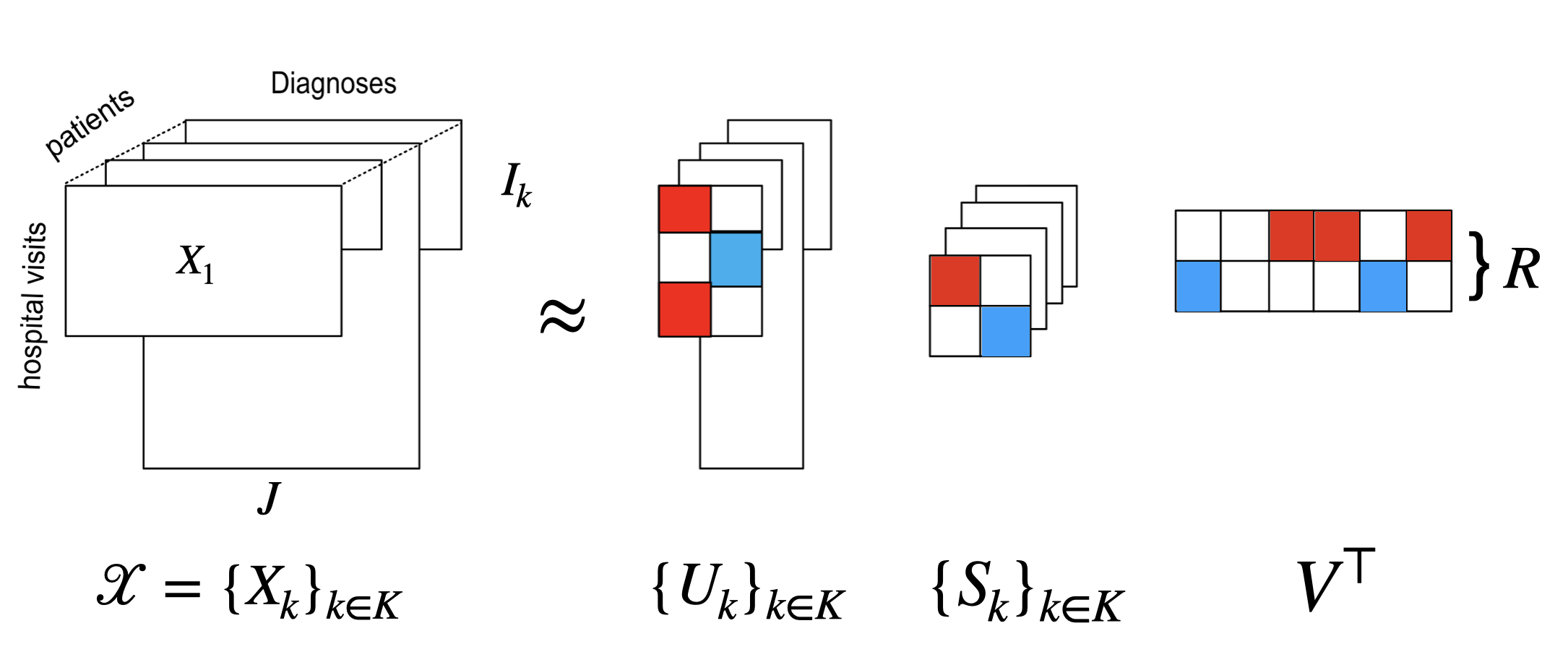}
    \caption{An example of the PARAFAC2 framework for temporal phenotyping. The input is a set of matrices $X_k$, where each matrix has $I_k$ rows (the number of visits for patient $k$) and $J$ columns (the shared clinical features, e.g., diagnoses). All patients use the same $J$ features but may have different visit counts $I_k$. The number of phenotypes corresponds to the rank $R$.}
    \label{fig:parafac_eg}
\end{figure}

The PARAFAC2 model is a more flexible variant of CP factorization proposed for modeling irregular tensors. Specifically, it maps each slice of an irregular tensor into a set of factor matrices. The estimation of the factor matrices in PARAFAC2 is often formulated as the following optimization problem:
\begin{equation}
    \begin{aligned}
    \underset{\{U_k\}, \{S_k\}, V}{\min} &\sum_{k=1}^{K} \frac{1}{2} \|X_k - U_k S_k V^\top\|_F^2, \\
    & \text {s.t} \quad U_k = Q_kH, \quad Q_k^\top Q_k = I, 
    \end{aligned}
\end{equation}
which solves for the factor matrix $U_k \in \mathbb R^{I_k \times R}$, the diagonal matrix $S_k$, and the invariant factor matrix $V$. Fig.~\ref{fig:parafac_eg} illustrates the PARAFAC2 factorization. The constraint, introduced by \citet{Hars1972b}, is imposed to ensure uniqueness of the decomposition. It is originally defined as $U_k^\top U_k = \Phi$, where \(\Phi\in\mathbb{R}^{R\times R}\) is a fixed but unknown matrix that is fixed across all slices \(k\). It can be equivalently expressed using column-wise orthogonality as $U_k = Q_kH $, where $Q_k^\top Q_k = I$ and $H\in\mathbb{R}^{R\times R}$ is an invariant matrix. The matrix \(H\) is learned by the PARAFAC2 algorithm. 

When applying PARAFAC2 to temporal EHR data, the decomposition results have the following interpretations:
\begin{enumerate}
  \item Each ${U}_k\in\mathbb{R}^{I_k\times R}$ provides the \textit{temporal trajectory} of each patient.  The $r$-th column of ${U}_k$ reflects the evolution of the expression of the $r$-th phenotype over all $I_k$ clinical visits.
  \item The diagonal matrix ${S}_k\in\mathbb{R}^{R\times R}$ denotes the relationship between the $k$-th patient and the set of phenotypes. Each column of ${S}_k$ corresponds to a phenotype, and if a patient has the highest weight in a specific column, then they are primarily associated with that particular phenotype \cite{Ren2020Tensor}.
  \item The common factor matrix ${V}\in\mathbb{R}^{J\times R}$ reflects the phenotypes and is common to all patients.  The non-zero values of the $r$-th column of ${V}$ denote the membership of the corresponding medical features to the $r$-th phenotype.
\end{enumerate}

In the context of EHRs, \(U_k S_k\) captures the \textit{phenotyping scores across visits for patient \(k\)}, while \(V\) encodes the membership of observed features in phenotypes.

\subsection{Dynamic Bayesian Network and Graph Notation}

We review the dynamic Bayesian networks (DBNs) and the associated graph notations. A dynamic Bayesian network (DBN) comprises an intra-slice weighted directed acyclic graph (static structure) that encodes dependencies within each time step, and an inter-slice weighted DAG (temporal structure) that encodes dependencies across successive time steps and is replicated between every pair of slices when the network is unrolled. A static structure is defined as an ordered pair $ G = (V_G, E_G)$, where \(V_G=\{1, 2, \dots, D\}\) denotes the set of nodes and $E_G = \{\,e_{ij} \mid i \to j,\; i,j\in V_G,\; i\neq j\}$ denotes the set of directed edges (or simply, edges). We say node $i$ is a parent of node $j$, denoted as $i \in \mathrm{pa}(j)$, where $\mathrm{pa}(j)$ is the set of all parents of $j$. To model temporal structures, we extend this definition by introducing a time-indexed edge set $E_G^{(p)} = \{\,e^{(p)}_{ki} \mid k \to i,\;k \in V^{t-p},\;i \in V_G^{t}\}$, which captures dependencies from time slice \(t-p\) to time slice \(t\) ($p  = 1,2,\ldots, P$), denotes the time‐lag order. In this temporal setting, node $k$ is considered a parent of node $i$ with a lag of $p$ time steps, denoted by $k \in \mathrm{pa}^{(p)}(i)$. We assume that the node set is identical across all time steps, i.e.\ \(V_G^{t}=V_G\) for every \(t\). Alternatively, both the static and temporal graph can be represented as adjacency matrices. We define the weighted intra-slice graph and weighted inter-slice graphs as: 
\begin{equation*}
W_{ij} =
\begin{cases}
w_{ij}, & e_{ij}\in E_G,\\
0,        & \text{otherwise.}
\end{cases}
\quad A^{(p)}_{ij} = 
\begin{cases}
a^{(p)}_{ij}, & e^{(p)}_{ki}\in E_G^{(p)},\\
0,        & \text{otherwise.}   
\end{cases}
\end{equation*}
Here, $w_{ij}$ and $a^{(p)}_{ij}$ are the edge weights, and $W,\{A^{(p)}\} \in \mathbb R^{D\times D}$ . Given the temporal observations \(\{x^{(t)}\}_{t=0}^{T}\), where \(x^{(t)} \in \mathbb{R}^D \), we can have the following formulation: 
\[
x_i^{(t)} = \sum_{j\in pa(i)} w_{ji}x_j^{(t)}+ \sum_{p=1}^P\sum_{k \in pa^{(p)}(i)} a^{(p)}_{ki} x_k^{(t-p)} +\epsilon_i^t,
\]
where $\epsilon_t\sim\mathcal{N}(0,1)$ is the noise term. Since the causal structure is a directed acyclic graph (DAG), the learning task is therefore to estimate an acyclic intra-slice graph and inter-slice graphs. Each \(A^{(p)}\) is automatically acyclic because edges only point forward in time (\(V^{t-p} \!\to\! V^{t}\)), prohibiting feedback loops from future to past. To enforce acyclicity, we use the constraint $h(W) = \mathrm{tr}\left( e^{W \circ W} \right) - d$, proposed by \citet{Zheng2018}. The problem can be solved via a continuous optimization with a score function $Sc(W,\{A^{(p)}\};D)$ as: 
\begin{equation}
\begin{split}
   &\min_{W,\{A^{(p)}\}} Sc(W, \{A^{(p)}\};D), \\
    & \text{s.t.} \quad h(W) = 0.  
\end{split}
\end{equation}
Since this is a pure data-driven approach, the $W$ and $\{A^{(p)}\}$ matrices are assumed to lie in a Markov Equivalence Class (MEC) \citep{Chang2024CDTD}. However, for EHR data, this formulation is inadequate. More precisely, an EHR system records $J$ biological features (e.g., diagnoses codes) at $I_k$ irregular visit times. When we run causal discovery algorithms directly on these raw data, we obtain a diagnosis-level causal graph (e.g., $W_{\text{diagnoses}} \in \mathbb R^{J\times J}$), rather than the casual graph (e.g., $W_{\text{phenotypes}} \in \mathbb R^{R\times R}$) among the clinically meaningful cluster (e.g., $R$ phenotypes).

\subsection{Causal Structure Among Latent Clusters}

\begin{figure}[H]
    \centering
    \includegraphics[width=0.8\linewidth]{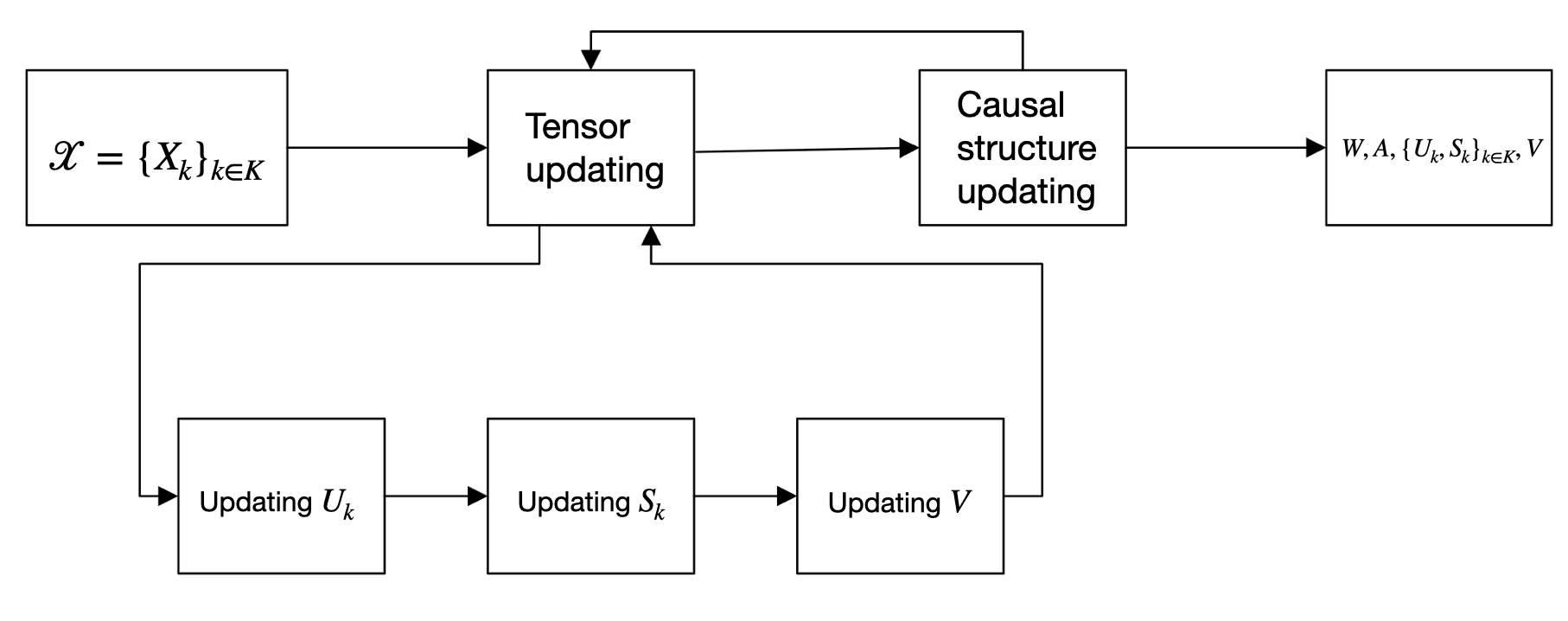}
    \caption{An overview of the proposed \texttt{CaRTeD} framework for causal phenotype network and computational phenotype.}
    \label{fig:method_fig}
\end{figure}

Next, we introduce our proposed framework, Causal Representation learning with Irregular Tensor Decomposition (CaRTeD), with a motivating example of learning the causal phenotype network. The key challenge addressed by CaRTeD is the integration of the temporal causal structure learning with the tensor decomposition. In the context of EHR data, we represent phenotype trajectories (or clusters in other settings) as $\tilde U_k = U_k S_k \in \mathbb{R}^{I_k \times R}$, and we assume each column contains observations across time for a single variable. In our problem, we assume a shared causal structure across slices and then model the temporal dynamics among latent phenotypes for $t \in \{p, p+1,\ldots, I_k\}$ as:

\begin{equation}
\label{eqn:causal}
\begin{split}
&\tilde u_{k_i}^{(t)} = \sum_{j\in pa(i)} w_{ji}\tilde u_{k_j}^{(t)}+ \sum_{p =1}^P\sum_{k \in pa^{(p)}(i)} a^{p}_{ki} \tilde {u}_{k_k}^{(t-p)} +\epsilon_{k_i}^t, 	
\\&\implies \tilde{U}_k = \tilde{U}_k W + \sum_{p = 1}^{P} \tilde{U}_k^{(i)} A^{(p)} + \epsilon_t,
\end{split}
\end{equation}
where $u_{k_i},u_{k_j},u_{k_k}$ is the $i$-th,$j$-th, $k$-th column of the $\tilde U_k$, respectively; the \(W\), \(\{A^{(p)}\}\), and \(\epsilon_t\) are as defined above; and $\tilde{U}_k^{(i)}$ is the time-lagged version of $\tilde{U}_k$ (See Section \ref{Methodology} for details on constructing the time-lagged version). Our goal is to estimate \(W\) and \(\{A^{(p)}\}\). However, the temporal phenotyping scores ($\tilde{U}_k$) are hidden variables and should be estimated through the PARAFAC2 decomposition. Therefore, To find the best causal structure among all slices, we consider the following separetable objective function that uses the ordinary least squares: 

\begin{equation}
\label{eq:pro_ori}
\begin{split}
&\min_{\substack{\{U_k\}, \{S_k\}, V, \\ W, \{A^{(p)}\}}} \mathcal{L}_{\text{PARAFAC2}} +\mathcal{L}_{\text{Causal}}\\
& = \sum_{k=1}^K \ \frac{1}{2} \|{X}_k - {U}_k{S}_k{V}^\top \|_F^2 + \frac{1}{2I_k}\| {U}_k{S}_k - {U}_k{S}_k W - \sum_{p = 1}^{P} {U}^{I_k-i}_k{S}_kA^{(p)} \|_F^2\\ & + \lambda_W \|W\|_1 + \lambda_A \sum_{p=1}^P\|A^{(p)}\|_1,\\
& \text{s.t.}  \quad U_k = Q_kH, \quad Q_k^\top Q_k = I,\quad \text{W is acyclic},
\end{split}
\end{equation}
where \(\ell_1 \)-norm penalties are incorporated to encourage sparsity. \textbf{This problem is not directly solvable} for several reasons. First, we have no prior information about these parameters (e.g., $W$, $U_k$, etc). Second, the formulation is non-convex because it contains multilinear terms. Even if we treat it as a tensor-decomposition problem with regularization, the second term is not only completely unknown but is also bilinear function. Hence, we solve the problem via block-coordinate descent (BCD) methods, whose key idea is to update each block iteratively. We demonstrate our methodology in Fig.~\ref{fig:method_fig}. Compared to learning the phenotype and causal diagram separately, our method not only provides a causally informed tensor decomposition with a novel regularization approach but also reduces the risk of suboptimal or incorrect causal structure due to estimation error, especially on small datasets (e.g., a small set of patients). For the sake of notation simplification, we define $A \coloneqq [{A^{(1)}}^\top|\ldots|{A^{(P)}}^\top]^\top$ by vertically concatenating of the transposed lag matrices. We will use this abbreviated notation and the full notation alternatively. 

\section{Methodology}
\label{Methodology}

Our \texttt{CaRTeD} is designed to jointly learn phenotypes and temporal causal phenotype networks from the irregular tensor data. One key challenge in this framework is the absence of any information about those parameters (e.g., $U_k$, $W$). More precisely, we cannot directly solve $W$ and $\{A^{(p)}\}$ since it is depended on the $U_k$ and $S_k$, which are obtained by the tensor decomposition, and vice versa. To address this, we propose a block-wise alternating minimization method to solve Eq.\eqref{eq:pro_ori}. In each iteration, we first update $\{U_k,S_k, V\}$ while keeping $W$ and $\{A^{(p)}\}$ fixed; then update $W$ and $\{A^{(p)}\}$ based on the updated factor matrices $\{U_k,S_k, V\}$. This iterative approach enables our framework to effectively perform tensor factorization under unknown or dynamically changing constraints.

\subsection{Updating the PARAFAC2 block}

To derive the update rule for the PARAFAC block, note that when updating $\{U_k, S_k, V\}$ with $W$ and $\{A^{(p)}\}$ fixed, the causal term acts as a regularization on $U_k$ and $S_k$. Existing PARAFAC2-based methods \citep{Perros2017SPARTan, Afshar2018COPA, Elif2025ssp, Zhao02072024} demonstrate that incorporating such constraints or regularizers often improves both performance and interpretability. This motivates our joint‐learning framework. Accordingly, we reformulate the subproblem as a regularized least‐squares problem. We introduce the following notation:

\[
f_{S_k}(U_k) = f_{U_k}(S_k) = f(U_k,S_k) = \frac{1}{2I_k}\| U_kS_k - U_kS_k W - \sum_{p = 1}^{P} U_k^{I_k-i}S_kA^{(p)} \|_F^2.
\]
For updating the PARAFAC2 factorization block, the problem Eq.\eqref{eq:pro_ori} can be rewritten as:
\begin{equation}
\label{eq:par_ori}
    \begin{split}
        &\min_{U_k,S_k,V}\sum_{k=1}^K \ \frac{1}{2} \|{X}_k - {U}_k{S}_k{V}^\top \|_F^2 + f(U_k,S_k),  \\
        & \text{s.t.} \quad U_k = Q_kH, \quad Q_k^\top Q_k = I,
    \end{split}
\end{equation}
where $H \in \mathbb R^{R\times R}$ is invariant with respect to $k$. The feasible set can be written as $\mathbb{S} = \{U_k \mid {U}_{k_1}^\top{U}_{k_1} = {U}_{k_2}^\top{U}_{k_2} = H^\top H, \quad k_1,k_2 \in [K]\coloneqq\{1,\ldots,K\}\}$ \citep{Hars1972b}. However, the causal information is regularized on two tensor components in a bilinear manner. In this case, we employ an Alternating Optimization (AO) for solving the PARAFAC2 in a block-wise manner as well. Then, the causal term is valid as a constraint here because it acts as a smooth regularizer on $U_k,S_k$. Note that this subproblem solves for $U_k, S_k, V$. For solving each block, we employ a consensus alternating direction method of multipliers (ADMM) scheme, which splits the problem into multiple subproblems, each solved approximately. Note that the optimization for each block is indeed convex. In our framework, we first solve $U_k$, then $S_k$ and $V$. 

\paragraph{Updating $U_k$ block:}
To solve the $U_k$ block, we fix all other variables ($S_k, V, W, A$) and obtain the following subproblem:
\begin{equation}
\label{U_k ori}
    \begin{split}
        &\min_{\{U_k\}_{k\leq K}}  \sum_{k=1}^{K} \|X_k - U_kS_kV^\top\|_F^2 + f_{{S}_k} ({U}_k), \\
        & \text{s.t} \quad U_k \in \mathbb{S}.
    \end{split}
\end{equation}
Because the feasible set is a non-convex, we rewrite the Eq.\eqref{U_k ori} into a standard form which is solvable by ADMM as follows:

\[
\min_{\{U_k\}_{k\leq K}} \sum_{k=1}^{K} \|X_k - U_kS_kV^\top\|_F^2 + f_{{S}_k} ( {U}_k) +  \iota_{S} \left( \{ {U}_k \}_{k\leq K} \right),
\]
where $\iota_{S}$ is the indicator function defined such that $\iota_{S} = 0$ if $ \{{U}_k \}_{k\leq K} \in \mathbb S$, and $\infty$ otherwise. For splitting the regularization of causal structure and PARAFAC2 constraints, we introduce two auxiliary variables $\tilde{U}_k, \hat{U}_k$, and formulate the following problem: 

\begin{equation}
    \begin{split}
    \min_{\{U_k,\tilde{U}_k,\hat{U}_k\}_{k\leq K}} &\sum_{k=1}^{K} \|X_k - U_kS_kV^\top\|_F^2 + f_{{S}_k} ( \tilde{U}_k) +  \iota_{S} \left( \{ \hat{U}_k \}_{k\leq K} \right), \\ 
    &\text{s.t} \quad U_k = \tilde{U}_k, \\
    & \qquad U_k = \hat{U}_k, \quad \forall k \in [K].   
    \end{split}
\end{equation}
As it is typical in the ADMM setting, we adopt the augmented Lagrangian method to solve the above constrained optimization problem. The augmented Lagrangian is a classical technique that converts a constrained problem into a sequence of unconstrained ones. We can then write the augmented Lagrangian as:

\begin{equation}
    \begin{split}
    \min_{\{U_k,\tilde{U}_k,\hat{U}_k\}_{k\leq K}} \sum_{k=1}^{K} &\|X_k - U_kS_kV^\top\|_F^2 + f_{{S}_k} ( \tilde{U}_k) +  \iota_{S} \left( \{ \hat{U}_k \}_{k\leq K} \right)\\ +
    &\frac{\rho_{u_k}}{2} \left\| U_k - \tilde{U}_k +\mu_{\tilde{U}_k}  \right\|_F^2 +
     \frac{\rho_{u_k}}{2} \left\| U_k - \hat{U}_k + \mu_{\hat{U}_k}\right\|_F^2, 
    \end{split}
\end{equation}
where $\rho_{u_k}$ is the penalty coefficient and $\mu_{\tilde{U}_k}, \mu_{\hat{U}_k} \in \mathbb{R}^{I_k \times R}$ are the Lagrange multipliers for each $k \in [K]$. 
Note that we use the scaled version of the augmented Lagrangian here. In this formulation, we update three variables $U_k, \tilde{U}_k, \text{and}\ \hat{U}_k$. We then have the following update rules. \textbf{To update $U_k$}, we solve the following problem: 

\begin{equation}
\label{U_k_ori}
\begin{split}
    U_k^{(t+1)} = \arg\min_{U_k}&  \|X_k - U_kS_kV^\top\|_F^2  + \frac{\rho_{u_k}}{2} \left\| U_k - \tilde{U}^{(t)}_k +\mu^{(t)}_{\tilde{U}_k}  \right\|_F^2 \\& + \frac{\rho_{u_k}}{2} \left\| U_k - \hat{U}_k^{(t)} + \mu^{(t)}_{\hat{U}_k}\right\|_F^2. 
\end{split}  
\end{equation}
Using the optimality condition, one obtains the closed-form update for \(U_k^{(t+1)}\). The detailed derivation is provided in Supplementary Material (see Section \ref{Up_U}).

\begin{equation}
\boxed{
U_k^{(t+1)} = \left(X_k V S_k^\top + \frac{\rho_{u_k}}{2} \left( \tilde{U}_k^{(t)} + \hat{U}_k^{(t)} - \mu^{(t)}_{\tilde{U}_k} - \mu^{(t)}_{\hat{U}_k} \right) \right) \left( S_k V^\top V S_k^\top + \rho_{u_k} I \right)^{-1}.
}   
\end{equation}
\textbf{To update $\tilde{U}_k$}, the following problem should be solved:

\begin{equation}
\label{U_t}
    \tilde{U}_k^{(t+1)} =\arg\min_{\tilde{U}_k} f_{{S}_k} ( \tilde{U}_k) + \frac{\rho_{u_k}}{2} \left\| U^{(t+1)}_k - \tilde{U}_k +\mu^{(t)}_{\tilde{U}_k}  \right\|_F^2.
\end{equation}
This problem cannot be solved directly since $f_{U_k} (\tilde{U}_k) = \frac{1}{2I_k}\| \tilde{U}_kS_k - \tilde{U}_kS_k W - \sum_{p = 1}^{P} \tilde{U}_k^{I_k-i}S_kA^{(p)} \|_F^2$ involves a time-lagged version of $U_k$. To ensure the mathematical consistency, we parametrize this formulation by using a shift matrix. The parametrized form is $\tilde{U}_k^{I_k-i} = M_i\tilde{U}_k = [0_i,I]^\top\tilde{U}_k$, where $M = [0_i,I]^\top$ and $0_i \in \mathbb{R}^{i \times I_k}$ is a zero vector or matrix with $i$ rows, corresponding to the autoregression order $p$. Thus, the problem can be rewritten as: 

\begin{equation}
    \tilde{U}_k^{(t+1)} =\arg\min_{\tilde{U}_k}  \frac{1}{2I_k}\| \tilde{U}S_k - \tilde{U}S_k W - \sum_{p = 1}^{P} M_i\tilde{U}S_kA^{(p)} \|_F^2 + \frac{\rho_{u_k}}{2} \left\| U^{(t+1)}_k - \tilde{U}_k +\mu^{(t)}_{\tilde{U}_k}  \right\|_F^2.
\end{equation}
To solve this, we vectorize the problem using the Kronecker product as follows: 

\[
\tilde{\mathbf{u}}_k = \arg\min_{\tilde{\mathbf{u}}_k} \frac{1}{2\,I_k}\;\bigl\|\Phi\,\tilde{\mathbf{u}}_k\bigr\|_2^2
\;+\;
\frac{\rho_{u_k}}{2}\,\bigl\|\mathbf{u}_k - \tilde{\mathbf{u}}_k\bigr\|_2^2,
\]
where $\Phi =(I-W)^\top S^\top \otimes I- \sum_{i=1}^p {A^{(p)}}^\top S_k^\top \otimes M_i$, $\mathbf{u}_k = \text{vec}\bigl(U_k^{(t+1)} + \mu_{\tilde{U}_k}^{(t)}\bigr)
$, and $\tilde{\mathbf{u}}_k = \text{vec}(\tilde{U}_k)$. Similarly, the closed-form of $\tilde{U}_k$ can be derived as: 

\[
\boxed{
\tilde{U}^{(t+1)}_k
\;=\;
mat\bigl[\Bigl(\frac{1}{I_k}\,\Phi^\top \Phi \;+\;\rho_{u_k}\,I\Bigr)^{-1}\,\rho_{u_k}\,\mathbf{u}_k\bigr].}
\]
We note that \textit{mat} is the de-vectorization operator that reshapes a vector back into its matrix form. For the full procedure of vectorizing the problem and solving the closed form, we include it in the supplementary material (see \S\ref{Up_tU_k}). \textbf{To update $\hat{U}_k$}, solve the following optimization problem:

\begin{equation}
   \hat{U}_k^{(t+1)} = \arg\min_{\hat{U}_k}   \iota_{S} \left( \{ \hat{U}_k \}_{k\leq K} \right) + \sum_{k=1}^{K}\frac{\rho_{u_k}}{2} \left\| U_k^{(t+1)} - \hat{U}_k +\mu^{(t)}_{\hat{U}_k}\right\|_F^2. 
\label{U_h}
\end{equation}
For evaluating Eq.\eqref{U_h}, it is equivalent to the projection onto $\mathbb{S}$. Therefore, we set $\hat{U}_k = Q_kH$ such that $Q_k^\top Q_k = I$, and solve the following problem:

\begin{equation}
\begin{split}
    \min_{H, \{Q_k\}_{k\leq K}} &\sum_{k = 1}^{K} 
     \frac{\rho_{u_k}}{2} \left\| U_k^{(t+1)} - Q_kH + \mu^{(t)}_{\hat{U}_k}\right\|_F^2, \\
     & \text{s.t.}\quad  Q_k^\top Q_k = I, \quad \forall k \in [K]. 
\end{split}
\end{equation}
We can observe that this problem needs to be solved in a block-wise manner as well. Fortunately, this problem can be solved efficiently. To update $Q_k$, we pose it as an individual Orthogonal Procrustes Problem \citep{Schnemann1966AGS} and solved by applying truncated SVD to $(U_k^{(t+1)} + \mu_k^{(t)})H^\top$. The closed-form solution is given by: 

\begin{equation}
\boxed{
    Q^{(t+1)}_k = U^{k}_{svd} (V^{k}_{svd})^\top,}
\end{equation}
where $U^{k}_{svd},\ (V^{k}_{svd})^\top$ are the  components of $U^{k}_{svd} \Sigma^k (V^{k}_{svd})^\top$. Then, we can derive a closed-form update for $H$ by setting the gradient of the objective function with respect to \( H \) to zero as an optimality condition. The full procedure is provided in the supplementary material (see \S\ref{Up_H}).

\begin{equation}
\label{eqn:H}
\boxed{
  H^{(t+1)} = \frac{1}{\sum_{k=1}^K \rho_{u_k}} \sum_{k=1}^K \rho_{u_k} Q_k^\top \left( U_k^{(t+1)} + \mu_{\hat{U}_k}^{(t)} \right).  }
\end{equation}
Finally, to update the dual variables, we use the following updates:

\begin{equation}
\label{U_dual}
    \begin{split}
        &\mu^{(t+1)}_{\tilde{U}_k} = \mu^{(t)}_{\tilde{U}_k} + U_k^{(t+1)} - \tilde{U}_k^{(t+1)},\\
        &\mu^{(t+1)}_{\hat{U}_k} = \mu^{(t)}_{\hat{U}_k} + U_k^{(t+1)} - \hat{U}_k^{(t+1)}.
    \end{split}
\end{equation}
In our algorithm, we follow the update order of $\hat U_k$, $\tilde U_k$, and $U_k$, and the update rules can be summarized in the Algorithm~\ref{alg:U_k}. We adopt the stopping criterion from \citet{Roald2022Aoadmm} for all tensor blocks, including the $S_k$ update.

\begin{algorithm}[H]
\caption{Updating of $U_k$ Block}
\label{alg:U_k}
\begin{algorithmic}
\STATE \textbf{Result:} $\{U_k\}_{k \leq K}$
\WHILE{stopping rule is not satisfied} 
\FOR{$k = 1, 2, \ldots, K$}
\STATE Update the $Q_k,H$ by solving the problem \eqref{U_h}.
\STATE Update the $\tilde{U}_k$ by solving the problem \eqref{U_t}.
\STATE Update the $U_k$ by solving the problem \eqref{U_k_ori}.
\STATE  Update the dual variables by solving the problem \eqref{U_dual}.
\ENDFOR
\ENDWHILE
\end{algorithmic}
\end{algorithm}

\paragraph{Updating $S_k$ and $V$:}
After updating $U_k$, we update the $S_k$ and $V$. To update $S_k$, we solve the following optimization problem involving an auxiliary variable $\tilde S$.
\begin{equation}
    \begin{split}
        &\min_{\{S_k\}_{k\leq K}}  \sum_{k=1}^{K} \|X_k - U_kS_kV^\top\|_F^2 + f_{{U}_k} (\tilde{S}_k), \\
        & \text{s.t} \quad S_k = \tilde{S}_k.
    \end{split}
\end{equation}
We can then write the augmented Lagrangian as:

\begin{equation}
    \begin{split}
    \min_{\{S_k,\tilde{S}_k\}_{k\leq K}} \sum_{k=1}^{K} &\|X_k - U_kS_kV^\top\|_F^2 + f_{{S}_k} ( \tilde{S}_k)+\frac{\rho_{s_k}}{2} \left\| S_k - \tilde{S}_k +\mu_{\tilde{S}_k}  \right\|_F^2.
    \end{split}
\end{equation}
To solve this problem, the main procedure is the same as that for solving the $U_k$ block. Hence, we omit the full procedure from the main text. The only difference is that $S_k$ is a diagonal matrix in this problem. Therefore, the vectorized form can be derived using the identity \(\text{vec}(U_k S_k V^\top) = (V \odot U_k) \text{vec}(S_k)\) as follows:

\begin{equation*}
    \left\| X_k - U_k S_k V^\top \right\|_F^2 = \left\| \mathbf{x}_k - (V \odot U_k) \mathbf{s}_k \right\|_2^2, \quad \left\| S_k - \tilde{S}_k + \mu \right\|_F^2 = \left\| \mathbf{s}_k - (\tilde{\mathbf{s}}_k - \boldsymbol{\mu}) \right\|_2^2,
\end{equation*}
where \(\mathbf{x}_k = \text{vec}(X_k)\), \(\mathbf{s}_k = \text{vec}(S_k)\), \(\tilde{\mathbf{s}}_k = \text{vec}(\tilde{S}_k)\), and \(\boldsymbol{\mu} = \text{vec}(\mu_{S_k}^{(t)})\). \textbf{To update $S_k$}, the problem can be rewritten as:

\begin{equation}
   \min_{\mathbf{s}_k} \left\| \mathbf{x}_k - (V \odot U_k) \mathbf{s}_k \right\|_2^2 + \frac{\rho_{s_k}}{2} \left\| \mathbf{s}_k - (\tilde{\mathbf{s}}_k - \boldsymbol{\mu}) \right\|_2^2. 
\end{equation}
The closed form solution for $S_k$ (the full procedure in \S\ref{Up_S}) is 

\begin{equation}
\label{S_k ori}
\boxed{
  S_k^{(t+1)} = mat \bigr[ \left( V^\top V*  U_k^\top U_k + \frac{\rho_{s_k}}{2} I \right)^{-1} \left(\text{diag} (U_k^\top X_k V) + \frac{\rho_{s_k}}{2} (\tilde{\mathbf{s}}_k - \boldsymbol{\mu}) \right) \bigl].} 
\end{equation}
Note that $*$ represents Hadamard product, and diag($\cdot $) extracts the diagonal elements into a vector. \textbf{To update $\tilde{S}_k$}, we solve the following optimization problem:

\begin{equation}
\label{S_t}
    \tilde{S}_k^{(t+1)} =\arg\min_{\tilde{S}_k} \sum_{k=1}^{K} f_{{S}_k} ( \tilde{S}_k) + \frac{\rho_{s_k}}{2} \left\| S^{(t+1)}_k - \tilde{S}_k +\mu^{(t)}_{S_k}  \right\|_F^2.
\end{equation}
To solve this, we also solve the vectorized problem as follows:

\begin{equation}
   \tilde{S}_k^{(t+1)} =\arg\min_{\mathbf{\tilde s}_k} \frac{1}{2\,I_k}\|T_k\,\mathbf{\tilde s}_k \|_{2}^{2}+
\frac{\rho_k}{2}\,\|\mathbf{\tilde s}_k - (\mathbf s_k + \boldsymbol{\mu})\|_{2}^{2},
\end{equation}
where $T_k = \bigl(I\odot U_k\bigr) - \bigl(W^\top \odot U_k\bigr) - \sum_{i=1}^{p} \bigl({A^{(p)}}^\top\odot U_k^{I_k - i}\bigr)$. We obtain the closed-form solution as follows:

\begin{equation}
\boxed{
\tilde{S}^{(t+1)}_k = mat \bigr[
\Bigl(\tfrac{1}{I_k}\,T_k^{T}\,T_k + \rho_{s_k}\,I\Bigr)^{-1}
\bigl(\rho_{s_k}(\mathbf s_k + \boldsymbol{\mu}) \bigl) \bigl].
}    
\end{equation}
The full procedure (e.g., the vectorization and the closed-form analysis) is provided in Section~\ref{Up_tS}. For updating the dual variables, we have:

\begin{equation}
\label{S_k dual}
\mu^{(t+1)}_{{S}_k} = \mu^{(t)}_{{S}_k} + S_k^{(t+1)} - \tilde{S}_k^{(t+1)}.
\end{equation}
Thus, the updating procedure can be summarized in the Algorithm~\ref{alg:S_k}.

\begin{algorithm}[H]
\caption{Updating of $S_k$ Block}
\label{alg:S_k}
\begin{algorithmic}
\STATE \textbf{Result:} $\{S_k\}_{k \leq K}$
\WHILE{stopping rule is not satisfied}
\FOR{$k = 1, 2, \ldots,K$}
\STATE Update the $\tilde{S}_k$ by solving the problem \eqref{S_t}.
\STATE Update the $S_k$ by solving the problem \eqref{S_k ori}.
\STATE  Update the dual variables by solving the problem \eqref{S_k dual}.
\ENDFOR
\ENDWHILE
\end{algorithmic}
\end{algorithm}

To update $V$, we solve the optimization problem as follows: 

\begin{equation}
       V^{(t+1)}  = \arg\min_{V} \sum_{k=1}^{K} \|X_k - U_kS_kV^\top\|_F^2. 
\end{equation}

Since we do not have any constraints on $V$, updating rule is trivial using the optimality condition. The closed-form solution is given by:

\begin{equation}
\label{V_updating}
\boxed{\;
V^{(t+1)}
=\Bigl(\sum_{k=1}^{K} X_k^{\!\top}U_k S_k\Bigr)
     \Bigl(\sum_{k=1}^{K} S_k^{\!\top}U_k^{\!\top}U_k S_k\Bigr)^{-1}.
\;}  
\end{equation}

To select the penalty parameters $\rho_{u_k}$ and $\rho_{s_k}$ for each block, inspired by \citep{Huang2016FEpara, Schenker2020AFO}, we set them as follows:

\begin{equation}
    \begin{split}
&\rho_{u_k} = \frac{1}{R} \operatorname{Tr} \left( S_k V^\top V S_k \right), \\
&\rho_{s_k} = \frac{1}{R} \operatorname{Tr} \left( V^\top V \ast U_k^\top U_k \right).
    \end{split}
\end{equation}

\subsection{Updating the Temporal Causal Block}
As we have discussed in the previous section, the optimization problem for updating $W, \{A^{(p)}\}$ is depended on the $U_kS_k$. Specifically, after fixing $U_k$ and $S_k$, we solve for $W$ and $\{A^{(p)}\}$ without incorporating any informed regularization. In contrast, when updating $U_k$, $W$ and $\{A^{(p)}\}$ still provide the relevant causal information. In this case, we need to minimize the following objective function:

\[
f(W,A) = \sum_{k=1}^K\frac{1}{2I_k}\| U_kS_k - U_kS_k W - \sum_{p = 1}^{P} U_k^{I_k-i}S_kA^{(p)} \|_F^2.
\]
However, to update the temporal causal block, the key challenge is to integrate patients record information to obtain a patient-invariant causal network structure. To address it, $W, \{A^{(p)}\}_{i=1}^p$ can be solved as follows, using the two auxiliary variables $\tilde W_k, \tilde A_k$. To simplify notation, we write $\tilde A_k = [A_1^k,A_2^k, \ldots, A_p^k]$ and use the abbreviated notation for A.

\begin{align*}
\min_{\{\tilde W_k,\tilde A_k\}_{k \in k},W,A} &\sum_{k=1}^K f(\tilde W_k,\tilde A_k) + \lambda_W \| W \|_1 + \lambda_A \| A \|_1,\\
& \quad \tilde W_k = W, \quad \tilde A_k = A, \quad \forall k \in [K],\\
& \quad \text{subject to } h(W) = 0, 
\end{align*}
where $h(W) = \mathrm{tr}\left( e^{W \circ W} \right) - d = 0$ is the acyclicity constraint. The problem can be solved efficiently with an ADMM-based aggregation strategy, which accurately learns the causal structure across all patients \citep{chen2025fDBN}. To transform the constrained problem into a series of unconstrained subproblems, the problem employs the augmented Lagrangian method as follows:

\begin{equation}
    \begin{split}
    \label{causal upd}
\mathcal{L}\left( \{ \tilde W_k, \tilde A_k \}_{k =1}^K, W, A, \alpha, \{ \beta_k, \gamma_k \}_{k =1}^K; \rho_1, \rho_2 \right) &=
\sum_{k=1}^K \Bigg[ f(\tilde W_k, \tilde A_k) + \frac{\rho_2}{2} \| \tilde W_k - W + \beta_k \|_F^2 \\
& \quad \frac{\rho_2}{2} \| \tilde A_k - A + \gamma_k \|_F^2 \Bigg]+ \lambda_W \| W \|_1 + \lambda_A \| A \|_1  \\
& + \frac{\rho_1}{2} (h(W)+\alpha )^2,
\end{split}
\end{equation}
where \( \{ \beta_k \}_{k =1}^K \in \mathbb{R}^{d \times d} \), \( \{ \gamma_k \}_{k =1}^K \in \mathbb{R}^{pd \times d} \) and \( \alpha \in \mathbb{R} \) are estimates of the Lagrange multipliers; \( \rho_1 \) and \( \rho_2 \) are the penalty coefficients. To solve this problem, we first obtain the $\tilde W_k, \tilde A_k$ for each subject by solving the following optimization problem: 

\begin{equation}
\label{Local}
\begin{split}
    (\tilde W_k^{(t+1)}, \tilde A_k^{(t+1)}) = & \arg \min_{\tilde W_k, \tilde A_k}  f(\tilde W_k, \tilde A_k) + \frac{\rho_2}{2} \| \tilde W_k - W + \beta_k \|_F^2 +\frac{\rho_2}{2} \| \tilde A_k - A + \gamma_k \|_F^2.
\end{split}
\end{equation}
The optimization problem admits a straightforward closed‐form solution via the optimality conditions in a simply way, so we omit the full derivation. Then we aggregate all the information to learn a single $W$ and $A$ by solving the following optimization problem:
\begin{equation}
\label{Global}
\begin{split}
    (W^{(t+1)}, A^{(t+1)}) = 
\arg \min_{W, A} & \sum_{k=1}^K \Bigg[\frac{\rho_2}{2} \| \tilde W_k - W + \beta_k \|_F^2 + \frac{\rho_2}{2} \| \tilde A_k - A + \gamma_k \|_F^2\Bigg] \\
&  +\frac{\rho_1}{2} (h(W)+\alpha )^2+ \lambda_W \| W \|_1 + \lambda_A \| A \|_1 .
\end{split}
\end{equation}
We cannot obtained the closed form since $\nabla h(W) =
(e^{\,W\circ W})^{T}\circ2W$. Therefore, we use the first-order methods to solve the optimization problem. Lastly, we update the dual variables by the following: 
\begin{equation}
\label{para}
\begin{split}
\beta_k^{(t+1)} & = \beta_k^{(t)} + \tilde W_k^{(t+1)} - W^{(t+1)}, \quad \gamma_k^{(t+1)}  = \gamma_k^{(t)} + \tilde A_k^{(t+1)} - A^{(t+1)}, \\
\alpha^{(t+1)} & = \alpha^{(t)} +h(W^{(t+1)}), \quad \rho_1^{(t+1)}  = \phi_1 \rho_1^{(t)}, \quad \rho_2^{(t+1)}  = \phi_2 \rho_2^{(t)}.
\end{split}
\end{equation}
Here, $\phi_1$ and $\phi_2$ are hyperparameters that determine the growth rate of the coefficients $\rho_1$ and $\rho_2$. The updates for $W, A$ are summarized in Algorithm~\ref{alg:sp-dbnsl}. For the causal block, the algorithm stops when $h(W)\le10^{-8}$.

\begin{algorithm}[H]
\caption{Updating Temporal Causal Block}
\label{alg:sp-dbnsl}
\begin{algorithmic}
\STATE \textbf{Result:} $W,A$
\WHILE{stopping rule is not satisfied}
    \STATE Update \( \tilde W_1^{(t+1)}, \ldots, \tilde W_k^{(t+1)}, \tilde A_1^{(t+1)}, \ldots, \tilde A_k^{(t+1)} \) for all subjects in parallel by Eq.\eqref{Local}. 
    \STATE Update the $W,A$ by aggregating the \( \tilde W_k^{(t+1)},  \tilde A_k^{(t+1)}\) for all $k$ by Eq.\eqref{Global}.
    \STATE Update the dual parameters \( \alpha^{(t+1)}, \rho_1^{(t+1)}, \rho_2^{(t+1)} \),  \( \beta_k^{(t+1)}, \gamma_k^{(t+1)}\) by Eq.\eqref{para}.
\ENDWHILE
\end{algorithmic}
\end{algorithm}
Overall, we can summarize the entire methodology, \texttt{CaRTeD}, in Algorithm~\ref{alg:CaRTeD}. Our method consists of one outer loop containing two inner blocks. For the tensor block, we solve the subproblem using an additional alternating‐optimization (AO) step. To ensure the efficiency of the algorithm, we only apply the stopping criterion from \citet{Roald2022Aoadmm} at the algorithm level and do not enforce any stopping rule between the two blocks. 

\begin{algorithm}[H]
\caption{CaRTeD: Temporal Causal Discovery from Irregular Tensor}
\label{alg:CaRTeD}
\begin{algorithmic}[1]
\REQUIRE Initial parameters $U_k, \tilde{U}_k, Q_k, H, S_k, \tilde{S}_k, V, \mu_{\tilde{U}_k},\mu_{\hat{U}_k}, \mu_{S_k}, W, \{A^{(p)}\}$
\FOR{$t = 1, 2, \ldots$}
\STATE Update $U_k, \tilde{U}_k, Q_k, \mu_{\tilde{U}_k},\mu_{\hat{U}_k}$ and $H$ by the Algorithm \ref{alg:U_k}.
\STATE Update $S_k, \tilde{S}_k,\mu_{S_k}$ by the Algorithm \ref{alg:S_k}.
\STATE Update $V$ by the Eq.\eqref{V_updating}.
\STATE Update $W,\{A^{(p)}\}$ by the Algorithm \ref{alg:sp-dbnsl}.
\ENDFOR
\STATE \textbf{Result:} $\{U_k, S_k\}_{k \leq K}, V, W, \{A^{(p)}\}$
\end{algorithmic}
\end{algorithm}

\section{Theoretical Analysis}
\label{sec:TA}

In this section, we present our theoretical results. Since our model is optimized via block coordinate descent (BCD), we need to discuss the convergence of each block. To the best of our knowledge, there is no existing theoretical analysis of irregular tensor decomposition. To update the tensor‐factorization sub-block, we solve:

\begin{equation*}
\begin{aligned}
&\min_{\{U_k,S_k\}_{k\leq K},V}\sum_{k=1}^K \ \frac{1}{2} \|{X}_k - {U}_k{S}_k{V}^\top \|_F^2 + f(U_k,S_k)  \\
& \text{s.t.} \quad {U}_{k_1}^\top{U}_{k_1} = {U}_{k_2}^\top{U}_{k_2} \quad \forall k_1,k_2 \in [K].
\end{aligned}
\end{equation*}
This is a \textbf{nonconvex optimization problem with a nonconvex constraint}. In our method, we employ the alternating optimization to solve this block. To solve each inner block, we apply the alternating direction method of multipliers with a consensus formulation. It is known that by augmenting the objective with a strictly convex penalty, one can guarantee that the AO routine converges to a stationary point, assuming each block’s ADMM subproblem is solved exactly in the limit of infinitely many inner iterations (as in Proposition 2.7.1. of \citep{Tseng2001CBCD}). This is why many existing methods \citep{Roald2022Aoadmm,Zhao02072024} incorporate convex regularization into each block update to guarantee convergence. However, these methods lack any theoretical convergence analysis because the PARAFAC2 constraint on \(U_k\) is \textbf{nonconvex}. Therefore, \textbf{the main contribution of our work is provide a convergence analysis of the \(U_k\)–block update (Algorithm \ref{alg:U_k}).} In our method, we impose causal‐informed regularization on the \(U_k\) and \(S_k\) blocks. Since \(V\) has no structural constraints, we do not regularize it; nevertheless, we demonstrate in the experimental section how convergence can still be achieved. Moreover, we expect that any convex regularizer (e.g., a small ridge term) would suffice. To save space and improve the readability of the theoretical analysis, we will use the simplified notation in our proof: 

\begin{equation*}
\begin{aligned}
&f^u_k(U_k)\; = \;f^s_k(S_k) =f(U_k,S_k,V) = \frac{1}{2} \|{X}_k - {U}_k{S}_k{V}^\top \|_F^2,\\
& h(U_k) = h(S_k) = \frac{1}{2I_k}\| U_kS_k - U_kS_k W - \sum_{p = 1}^{P} U_k^{I_k-i}S_kA^{(p)} \|_F^2.
\end{aligned}
\end{equation*}
When we solve this problem in a \textbf{block-wise manner}, we can observe that the objective functions (e.g., $f^u_k(U_k), f^s_k(S_k)$) \textbf{are smooth and the causal regularization term is convex}. Furthermore, when we update the $S_k$ block, the sub‑problem can be viewed as a convex quadratic optimization with a smooth regularizer. When updating the $U_k$ block, the objective remains convex and smooth, but is solved under a nonconvex orthogonality constraint. In our theoretical analysis, we first analyze the updating rule for $S_k$ (Algorithm \ref{alg:S_k}), which has no constraint. Then, we analyze the updating rule for the $U_k$ block with an additional nonconvex constraint. Since the $U_k$ subproblem without the nonconvex constraint is analogous, we only provide detailed proofs of the key conclusions for $S_k$. Note that the convergence analysis is carried out with the standard (un-scaled) ADMM formulation, which is equivalent to the scaled version introduced in the methodology section. For a comprehensive review of ADMM, see \citet{Boyd2011ADMM}.

For the nonconvex constraint $\mathbb S = \bigl\{\,U_k \mid U_k = Q_k H,\;Q_k^\top Q_k = I,\;H\in\mathbb{R}^{n\times n}\bigr\},$ the pair \((Q_k,H)\) defines the feasible region for \(U_k\).  Updating \(Q_k\) reduces to an orthogonal Procrustes problem, which admits a unique closed-form solution via the SVD and converges in one step \citep{Kiers1999parafac}. By given the algorithm, we can see that $H^{(t)}$ is updated by using $U^{(t+1)}$ and $\mu^{(t)}$, which is a weighted linear combination. Since the map \((Q_k,H)\mapsto Q_kH\) is continuous and the set \(Q_k\) is compact, the property of updating \(H^{(t)}\) over $t$ is important for AO-ADMM convergence guarantee.

\subsection{Analysis of Algorithm \ref{alg:S_k}}

Before we begin the analysis of Algorithm \ref{alg:S_k}, we first present some useful lemmas. Note that these lemmas are analogous to those for Algorithm \ref{alg:U_k}, with proofs of the same form.

\begin{lemma}[Lipschitz gradient]
\label{lemma:Li-smooth}
For all \(i\in[K]\), each function \( f^s_i\) is \(L_i\)-smooth ($f^u_i$ as well), that is, for every \(x_i,\hat x_i\),
\[
\|\nabla f^s_i(x_i)-\nabla f^s_i(\hat x_i)\|\;\le\;L_i\,\|x_i-\hat x_i\|.
\]
As a consequence (cf.\ Lemma 1.2.3 in \cite{Nesterov2003}), we also have
\[
\bigl|f_i(x_i)-f_i(\hat x_i)-\langle\nabla f_i(\hat x_i),\,x_i-\hat x_i\rangle\bigr|
\;\le\;\frac{L_i}{2}\,\|x_i-\hat x_i\|^2.
\]
\end{lemma}

By using Lemma \ref{lemma:Li-smooth}, we have the following result.

\begin{lemma}
\label{lemma:smooth_result}
    In Algorithm \ref{alg:S_k}, we can have the following 
    \[
    L_k^2 \,\bigl\lVert S_k^{(t+1)} - S_k^{(t)}\bigr\rVert^2 \ge
    \bigl\lVert \mu^{(t+1)}_{\tilde{S}_k}  - \mu^{(t)}_{\tilde{S}_k} \bigr\rVert^2,
    \quad \forall\,k=1,\dots,K.
    \]
\end{lemma}

Next, we apply Lemma~\ref{lemma:smooth_result} to bound the change in the augmented Lagrangian resulting from the \(S_k\)-block update.

\begin{lemma}
\label{lemma:S_ bound}
For the updating rule, we have the following with the strong‑convexity modulus $\gamma_k(\rho_k),\tilde \gamma_k(\rho_k)$
\[
\mathcal L\bigl(\{S_k^{(t+1)}, \tilde S_k^{(t+1)}\},\mu^{(t+1)}_{\tilde{S}_k}\bigr) 
- 
\mathcal L\bigl(\{S_k^{(t)}, \tilde S_k^{(t)}\},\mu^{(t)}_{\tilde{S}_k}\bigr)
\leq \sum_{k=1}^{K}
(\frac{L_k^2}{\rho_k} -\frac{\gamma_k(\rho_k)}{2}) \|S_k^{(t+1)}-S_k^{(t)}\|^{2}- 
\frac{\tilde \gamma_k(\rho_k)}{2}\,
\bigl\lVert \tilde S_k^{(t+1)}-\tilde S_k^{(t)}
\bigr\rVert^{2}.
\]
\end{lemma}

In this case, we can always find a sufficiently large $\rho_k$ when $\gamma_k(\rho_k) \neq 0$ and $\tilde \gamma_k(\rho_k) = \gamma_0 $ such that $\rho_k\gamma_k(\rho_k) \geq 2L_k^2$. Consequently, the augmented Lagrangian function will always decrease. Thus, we show that $\mathcal L\bigl(\{S_k^{(t)}, \tilde S_k^{(t)}\},\mu^{(t)}_{\tilde{S}_k}\bigr)$ is convergent.

\begin{theorem}[Algorithm \ref{alg:S_k} is convergent]
\label{Thm:convegent}
Suppose each \(\rho_k\) is sufficiently large. Then the augmented‐Lagrangian sequence
\[
 \mathbf L^{(t)} = \mathcal L\bigl(\{S_k^{(t)},\,\tilde S_k^{(t)}\},\,\mu_{\tilde S_k}^{(t)}\bigr)
\]
is monotonically decreasing, bounded below by a finite constant, and therefore convergent:
\[
\lim_{t\to\infty} \mathbf L^{(t)} = \mathbf L^* > -\infty.
\]
\end{theorem}

To show that $\mathcal L(\{S_k^{(t)},\tilde S_k^{(t)}\},\mu_{\tilde S_k}^{(t)})$ converges to the set of stationary solutions, the statement can be proved using Theorem 2.4 in \cite{Hong2016}, since all of its assumptions and required properties have been verified for this problem. Therefore, we omit the formal proof. For analysis of the $U_k$-update, we can observe that the $U_k$-update only differs from the $S_k$-update by an additional constraint set. Therefore, the key goal in analyzing the \(U_k\)‐update is to show that \textit{the same objective converges to a stationary point under the imposed nonconvex constraint.}

\subsection{Analysis of Algorithm \ref{alg:U_k}}
We slightly change the problem formulation before presenting the proof. The optimization problem for $U_k$ is

\begin{equation}
\label{rewrite:uk}
\begin{aligned}
 &\min_{\{U_k\}_{k\leq K}}f_k^u(U_k) =  \sum_{k=1}^{K} \|X_k - U_kS_kV^\top\|_F^2 + h({U}_k), \\
& \text{s.t} \quad U_k \in \mathbb S,
\end{aligned}
\end{equation}
where $\mathbb S = \{U_k \mid U_k = Q_kH, \quad Q_k^\top Q_k = I, \quad H\in \mathbb R^{n\times n} \}$. To enforce this constraint, we incorporate the indicator function defined in the previous section. We then express the problem as a consensus‐form augmented Lagrangian, as done in our algorithm.

\begin{equation}
\label{rewrite:uk full}
\begin{aligned}
 \min_{\{U_k\}_{k\leq K}} \sum_k &f_k^u(U_k) + \iota_S (\hat U_k), \\
& \text{s.t} \quad U_k = \hat U_k.
\end{aligned}
\end{equation}
As we mentioned earlier, it is not hard to verify Lemma \ref{lemma:Li-smooth}, Lemma \ref{lemma:S_ bound}, and Theorem \ref{Thm:convegent} for $f_k^u(U_k)$. Note that we only use the optimality condition and strong convexity of augmented Lagrangian to show the previous lemma. To obtain the analogous result for $U_k$, the only modification is replacing the gradient term $\nabla_{\tilde U_k} \mathcal L$ with the subgradient $\partial_{\tilde U_k} \mathcal L$, because the $\tilde U_k$-update contains an indicator function and is therefore nonsmooth. Before proving the theorem, we introduce the following definition and Lemmas:

\begin{definition}[Coercivity over a feasible set]
\label{coercivity}
Let $\mathcal F\subseteq\mathbb R^{m}\times\mathbb R^{n}$ be a feasible set and  
let $\varphi:\mathbb R^{m}\times\mathbb R^{n}\to\mathbb R\cup\{+\infty\}$ be an extended–real‑valued objective function.  
We say that \textbf{$\varphi$ is coercive on $\mathcal F$} if, for every sequence
$\{(x_k,y_k)\}_{k\ge 1}\subseteq\mathcal F$ with
$\|(x_k,y_k)\|\to\infty$, we have
\[
\varphi(x_k,y_k)\;\longrightarrow\;+\infty.
\]
Equivalently,
\[
\|(x,y)\|\xrightarrow[(x,y)\in\mathcal F]{}\infty
\quad\Longrightarrow\quad
\varphi(x,y)\xrightarrow{}\infty.
\]
\end{definition}

\begin{lemma}[Bounded sequence]
\label{lemma:Uk bounded}
For all $k \in [K]$, suppose $\rho_k$ is large enough, then, the sequence ${(U^{t}_k,\,\hat{U}^{t}_k,\,\mu^{t}_{\hat{U}_k})\bigr\}_{t=0}^\infty}$ produced by Algorithm \ref{alg:U_k} satisfy:
\begin{enumerate}
  \item\emph{(Monotonicity)}:
    $\displaystyle 
      \mathcal{L}(U^{t}_k,\,\hat{U}^{t}_k,\,\mu^{t}_{\hat{U}_k})\ge \mathcal{L}(U^{t+1}_k,\,\hat{U}^{t+1}_k,\,\mu^{t+1}_{\hat{U}_k}).$
  \item\emph{(Lower‐boundedness)}:
    $\{\mathcal{L}(U^{t}_k,\,\hat{U}^{t}_k,\,\mu^{t}_{\hat{U}_k})\}_{t\in\mathbb N}$ is bounded below and hence converges as $t \to\infty$.
    \item \emph{(Boundedness)}:
    The sequence $\{U^{t}_k,\,\hat{U}^{t}_k,\,\mu^{t}_{\hat{U}_k}\}_{t\in\mathbb N}$ is bounded.
\end{enumerate}
\end{lemma}

\begin{lemma}[Subgradient bound]
\label{Lemma: Uk Subgradient bound}
There exists a constant $C(\rho)>0$  and $\|d^{t+1}\|\in \partial\mathcal{L}(U^{t+1},\hat U^{t+1},\mu_{\hat{U}_k}^{t+1})$ such that, 
\begin{equation}
    \begin{aligned}
        \|d^{(t+1)}\| \leq C(\rho) (\sum_{k} \| \hat U^{(t+1)}_{k} - \hat U^{(t)}_{k}\| + \| U^{(t+1)}_{k} -U^{(t)}_{k} \|)
    \end{aligned}
\end{equation}
\end{lemma}

\begin{lemma}[Limiting continuity]
\label{lemma: Uk limit cont}

If $(U_k^*,\hat U_k^*,\mu_{\hat U_k}^*)$ is the limit point of a subsequence $(U_k^{t_s},\hat U_k^{t_s},\mu_{\hat U_k}^{t_s})$ for $s\in\mathbb{N}$, then
\begin{equation} 
 \lim_{s\to\infty}\mathcal{L}(U_k^{t_s},\hat U_k^{t_s},\mu_{\hat U_k}^{t_s}) =  \mathcal{L}(U_k^*,\hat U_k^*,\mu_{\hat U_k}^*). 
\end{equation}
\end{lemma}

We have the following theorem for the convergence of $U_k$. 

\begin{theorem}
\label{Thm: Uk_con}
For any sufficiently large \(\rho_{u_k}\), the sequence 
\[
\bigl\{(U^{t}_k,\,\hat{U}^{t}_k,\,\mu^{t}_{\hat{U}_k})\bigr\}_{t=0}^\infty
\]
generated by Algorithm \ref{alg:U_k} has at least one limit point and each limit point is a stationary point of the augmented Lagrangian \(\mathcal L(U^{t}_k,\,\hat{U}^{t}_k,\,\mu^{t}_{\hat{U}_k})\). 
\end{theorem}

Our theoretical analysis not only demonstrates the convergence of the tensor decomposition block, but also bridges the gap in convergence guarantees for AO-ADMM based PARAFAC2 methods \citep{Roald2022Aoadmm, Huang2016FEpara}. More importantly, our analysis provides key insights for designing AO-ADMM based tensor decomposition frameworks, particularly regarding the boundedness under nonconvex constraints and the properties of the regularization functions. For updating the causal block, the convergence analysis has been proved by \citet{ng2022convergence}. Therefore, we omit theoretical analysis in our paper. Provided each block sub-problem has a unique minimizer and is solved exactly (i.e., given infinitely many inner ADMM iterations), both the causal block and the tensor-decomposition block reach their blockwise minima at every outer step. Under these standard AO conditions, the overall algorithm converges to a stationary point.

\section{Performance Evaluation Using Simulated Experiments}
In this section, we evaluate the performance of \texttt{CaRTeD} on simulated datasets generated from an irregular tensor with embedded causal effects. We benchmark our method against two baselines using six evaluation metrics, three for causal structure recovery and three for tensor factorization quality, to demonstrate model effectiveness.

\subsection{Data Generation and Settings}
We generate a synthetic irregular tensor \(\mathcal{X} = \left\{ {X}_k \in \mathbb{R}^{I_k \times J} \right\}_{k=1}^{K}\) by generating each of the ground-truth factors. Specifically, given the true rank $R$ and number of medical features $J$, we sample matrices \(H \in \mathbb{R}^{R \times R}\) and \(S_k \in \mathbb{R}^{R \times R}\) element-wise from a uniform distribution over the interval \([5, 10]\). The factor matrix \(V \in \mathbb{R}^{J \times R}\) is drawn from the same distribution. To better reflect the clustered structure of real-world phenotypes, we apply a post-processing step to enforce such a structure in \(V\), thereby enhancing the biological realism. Given the number of hospital visits $I_k$ for each patient, we generate $Q_k$, $\forall k \in \{1,\dots, K\}$, as a binary, non-negative matrix whose columns are orthonormal; that is, $Q_k^{\!\top}Q_k = I$
, and define \(U_k = Q_k H\). For the causal structure, we generate the intra-slice matrix \(W \in \mathbb{R}^{R \times R}\), which is a DAG, and inter-slice matrices \(A^{(p)} \in \mathbb{R}^{ R \times R }\) using Erdős--Rényi (ER) graphs, where $i = 1, \ldots, p$ and \(p\) denotes the autoregressive order. The causal effect on each  product \(U_k S_k\) is incorporated as described in Eq.\eqref{eqn:causal}. Details of the causal structure simulation are provided in the Supplementary Materials (see \S\ref{sim}). Finally, each slice of the irregular input tensor \(\left\{ {X}_k \in \mathbb{R}^{I_k \times J} \right\}_{k=1}^{K}\) is generated as:
\[
{X}_k = U_k S_k V^\top + \epsilon,
\]
where \(\epsilon\) represents noise. In our experiments, we use random initialization for \(U_k\), \(S_k\), and \(V\), and initialize \(W = I\) and \(A = 0\).

\subsection{Benchmark Methods}
As our method jointly learns both tensor decomposition and causal phenotype networks from an irregular tensor augmented with causal structure, we evaluate its performance from two complementary perspectives. From the tensor decomposition perspective,  we apply the original constrained PARAFAC2 model, COPA \citep{tensorly}, with non-negative constraint to the data to assess decomposition quality. Following the \texttt{PARAFAC2} demonstration by \texttt{TensorLy}, we fit ten models and select the best one. From the causal structure learning perspective, we gather phenotype information (e.g., the $U_kS_k$ factors) produced by COPA and then directly apply the existing DBN learning framework, denoted \texttt{DDBN}, to learn temporal causal networks. Unlike our proposed joint‐learning approach, these benchmark methods proceed in a separate and sequential manner, without feedback loops between decomposition and structure learning.

\subsection{Evaluation Metrics}
Since causal network inference and tensor decomposition address different aspects of our problem, we evaluate performance using metrics from both perspectives. From the tensor side, we assess how well the proposed method and benchmarks recover the true simulated phenotype factor matrix using similarity measures on the causal irregular tensor. Specifically, we compute the similarity (SIM) between the estimated phenotype matrix \(V_{\text{est}}\in\mathbb{R}^{J\times R}\) and the ground-truth phenotype matrix \(V\in\mathbb{R}^{J\times R}\). First, we define the cosine similarity between vectors \(\mathbf{v}_i\) and \(\hat{\mathbf{v}}_j\) as $C_{i,j} = \frac{\mathbf{v}_i^\top \hat{\mathbf{v}}_j}{\|\mathbf{v}_i\|\;\|\hat{\mathbf{v}}_j\|}.$ Then,
\[
\mathrm{SIM}(V, V_{\text{est}}) = \frac{1}{R} \sum_{i=1}^{R} \max_{1\le j\le R} C_{i,j}.
\]
SIM ranges from 0 to 1, with values closer to 1 indicating greater similarity. We also compute the cross-product invariance (CPI) to evaluate recovery of \(U_k\). CPI is defined as
\[
\mathrm{CPI} = 1 - \frac{\sum_{k=1}^{K}\bigl\lVert U_k^\top U_k - H^\top H\bigr\rVert_F^2}{\sum_{k=1}^{K}\bigl\lVert H^\top H\bigr\rVert_F^2},
\]
which can range from \(-\infty\) to 1; values near 1 indicate more accurate recovery of the underlying factors. Finally, treating \(U_kS_k\) as a temporal phenotype trajectory, we define the recovery rate (RR) of the estimated \(X_k^{(\text{est})}=U_kS_k\) relative to the ground-truth \(X_k\) as
\[
\mathrm{RR}
= 1 - \frac{\sum_{k=1}^{K}\bigl\lVert X_k^{(\text{est})\top}X_k^{(\text{est})} - X_k^\top X_k\bigr\rVert_F^2}
             {\sum_{k=1}^{K}\bigl\lVert X_k^\top X_k\bigr\rVert_F^2}.
\]
From the causal discovery side, we evaluate graph recovery using three metrics: Structural Hamming Distance (SHD), True Positive Rate (TPR), and False Discovery Rate (FDR). Mathematically, Given the $A^{true}$ and $A^{estimated}$, SHD is defined as
$$
\mathrm{SHD}
  = \sum_{i\neq j}\Bigl[\,
      \underbrace{\mathbf 1\{A^{true}_{ij}=1\land A_{ij}^{estimated}=0\}}_{\text{missing}}
      +\,
      \underbrace{\mathbf 1\{A^{true}_{ij}=0\land A_{ij}^{estimated}=1\}}_{\text{extra}}
      +\,
      \underbrace{\mathbf 1\{A^{true}_{ij}=1\land A_{ij}^{estimated}=1\}}_{\text{misoriented}}
    \Bigr].
$$
A true positive (TP) is an edge that is correctly recovered, a false positive (FP) is a spurious edge, and a false negative (FN) is a missed true edge. The true positive rate (TPR) and false discovery rate (FDR) are therefore

$$
\text{TPR}=\frac{\text{TP}}{\text{TP}+\text{FN}},\qquad
\text{FDR}=\frac{\text{FP}}{\text{TP}+\text{FP}}.
$$
SHD measures the dissimilarity between the inferred and true graphs by counting missing edges, extra edges, and incorrectly oriented edges; smaller SHD indicates better alignment. TPR (sensitivity or recall) is the ratio of true positives to the sum of true positives and false negatives; higher TPR indicates more true edges correctly identified. FDR is the ratio of false positives to the sum of false positives and true positives; lower FDR indicates fewer incorrect edges. Together, these metrics provide a comprehensive evaluation of the inferred causal structure.

\subsection{Experimental Results}
Our experiments focus on two complementary tasks, tensor decomposition and causal discovery, and are structured into two evaluation scenarios. In the first scenario, we assess tensor decomposition performance. We set the number of features to \(J = 12\), the number of slices to \(K = 100\), and the rank to \(R = 4\), drawing each \(I_k\) uniformly at random from the interval \([10,\,21]\). The data generation procedure for the second scenario is analogous, and more details are provided in a later section. To evaluate our method from the tensor decomposition perspective, we vary the noise level \(\epsilon \in \{0.1, 0.25, 0.5, 1.0\}\) and report the three recovery metrics described above. Since RR and CPI both range from \(-\infty\) to 1 (values closer to one are better), we denote negative values as \emph{NeN} to improve table readability. We evaluate our method over 20 replications for each noise level, and the results are shown in Table~\ref{tab:tensor_result}. For our \texttt{CaRTeD}, we present two types of results: one with a random start (\texttt{CaRTeD}) and one with a warm start (\texttt{W-CaRTeD}) with an approximated $\tilde V$, since we found that initializing \(\tilde V\) with a warm start yields better performance and computational efficiency. Note that initialization strategies, such as performing multiple runs, have been introduced by \citet{Roald2022Aoadmm} and are crucial in the AO setting. To approximate \(\tilde V\), we first perform several runs of pure tensor decomposition. Because real‐world phenotypes exhibit a clustered structure, we apply a small threshold to \(V\) as the off‐diagonal entries are relatively small. Moreover, since our method regularizes on \(U_k\) and \(S_k\), warm starts for these factors are not feasible as \(W\) and \(A\) are completely unknown. Following the hyperparameter-tuning suggestions by \citet{chen2025fDBN}, we set \(\lambda_W = \lambda_A = 0.5\) and apply thresholds of 0.3 and 0.1 to \(W\) and \(A\), respectively.

\begin{table}[H]
  \centering
  \caption{Comparison of tensor decomposition performance under different noise levels.}
  \label{tab:tensor_result}
  \begin{tabular}{l l c c c c c}
    \toprule
    Method   & Metric & 0.00 & 0.10 & 0.25 & 0.50 & 1.00 \\
    \midrule
    \multirow{3}{*}{W-CaRTeD}
     & CPI & $.761\pm.015$ & $.719\pm.019$ & $.719\pm.019$ & $.714\pm.019$ & $.709\pm.019$ \\
     & SIM & $.999\pm.000$ & $.999\pm.000$ & $.999\pm.001$ & $.999\pm.001$ & $.999\pm.001$ \\
     & RR  & $.981\pm.007$ & $.964\pm.010$ & $.964\pm.010$ & $.964\pm.010$ & $.964\pm.010$ \\
     \midrule
     \multirow{3}{*}{CaRTeD}
     & CPI & $.423\pm.036$ & $.398\pm.036$ & $.398\pm.036$ & $.398\pm.036$ & $.398\pm.036$ \\
     & SIM & $.931\pm.012$ & $.912\pm.015$ & $.912 \pm .015$ & $.912\pm.015$ & $.912\pm.015$ \\
     & RR  & $.612\pm.034$ & $.583\pm.064$ & $.583\pm.064$ & $.583\pm.064$ & $.583\pm.064$ \\
    \midrule
    \multirow{3}{*}{COPA}
    & CPI & $.022\pm.578$ & $.009\pm.626$ & $.041\pm.563$ & $.010\pm0.571$ & NeN \\
    & SIM & $.940\pm.015$ & $.938\pm.021$ & $.938 \pm .021$ & $.938\pm.020$ & $.938\pm.020$ \\
    & RR  & NeN & NeN & NeN & NeN & NeN \\
    \bottomrule
  \end{tabular}
\end{table}

From Table~\ref{tab:tensor_result}, we observe that the CPI for \texttt{COPA} is much lower than for both \texttt{W-CaRTeD} and \texttt{CaRTeD} in all cases. This is reasonable, since \texttt{COPA} enforces only a non-negativity constraint and does not incorporate any causal-structure information; we believe this lack of structure causes \texttt{COPA} to misinterpret during the decomposition. As the noise scale increases, the performance of our methods (\texttt{CaRTeD} and \texttt{W-CaRTeD}) remains stable, whereas the CPI for \texttt{COPA} becomes negative when $\epsilon = 1.0$. However, we can see that the SIM scores of \texttt{COPA} are about $0.1$–$0.2$ higher than those of \texttt{CaRTeD} across all noise levels, and the SIM values for both methods stabilize as noise increases. When we provide a warm-start $\tilde V$, \texttt{W-CaRTeD} delivers excellent results as SIM values reach approximately 0.99, which is about 0.05 higher than that of \texttt{COPA} when $\epsilon = 0$, and 0.06 better than that when the noise becomes larger. In contrast, \texttt{COPA} has negative RR values under all noise conditions, which is consistent with its lower CPI. In comparison, our methods yield reasonable RR scores. Besides the general comparison, we can more closely compare the results with and without the warm-start $\tilde V$. From the table, we can see both metrics improve, especially for the CPI and RR. More importantly, the results of RR are improved by 0.3 for all cases, which is consistent with the 0.3 improvement in CPI. These comparisons show the outstanding performance of our proposed methods.

\begin{table}[ht]
\centering
\caption{Recovery performance of the causal phenotype network for intra-slice network \(W\).}
\label{tab:causal_W}
\begin{tabular}{l l c c c c}
\toprule
Method & Metric & 10 & 20 & 40 & 80 \\
\midrule
\multirow{3}{*}{CaRTeD}
 & SHD & $3.000 \pm 0.000$ & $2.800 \pm 0.400$ & $2.400 \pm 0.490$ & $2.600 \pm 0.490$ \\
 & FDR & $0.250 \pm 0.000$ & $0.240 \pm 0.020$ & $0.220 \pm 0.024$ & $0.230 \pm 0.024$ \\
 & TPR & $0.600 \pm 0.000$ & $0.640 \pm 0.080$ & $0.720 \pm 0.098$ & $0.680 \pm 0.098$ \\
\midrule
\multirow{3}{*}{DDBN}
 & SHD & NA & $5.200 \pm 1.327$ & $5.800 \pm 0.748$ & $5.200 \pm 1.600$ \\
 & FDR & NA & $0.400 \pm 0.389$ & $0.567 \pm 0.327$ & $0.593 \pm 0.339$ \\
 & TPR & NA & $0.200 \pm 0.310$ & $0.120 \pm 0.098$ & $0.250 \pm 0.158$ \\
\bottomrule
\end{tabular}
\end{table}

\begin{table}[ht]
\centering
\caption{Recovery performance of the causal phenotype network for inter-slice network \(A\).}
\label{tab:causal_A}
\begin{tabular}{l l c c c c}
\toprule
Method & Metric & 10 & 20 & 40 & 80 \\
\midrule
\multirow{3}{*}{CaRTeD}
 & SHD & $8.500 \pm 0.500$ & $8.800 \pm 1.470$ & $10.00 \pm 0.800$ & $9.000 \pm 1.000$ \\
 & FDR & $0.714 \pm 0.018$ & $0.728 \pm 0.038$ & $0.731  \pm 0.038$ & $0.757 \pm 0.023$ \\
 & TPR & $0.750 \pm 0.000$ & $0.850 \pm 0.122$ & $0.875  \pm 0.125$ & $0.950 \pm 0.100$ \\
\midrule
\multirow{3}{*}{DDBN}
 & SHD & NA & $8.000\pm 2.500$ & $10.00 \pm 1.200$ & $10.00\pm 1.000$ \\
 & FDR & NA & $0.611 \pm 0.452$ & $0.769 \pm 0.063$ & $0.833 \pm 0.056$ \\
 & TPR & NA & $0.250 \pm 0.200$ & $0.312 \pm 0.325$ & $0.375 \pm 0.125$ \\
\bottomrule
\end{tabular}
\end{table}

To compare results in causal graph learning among the patients, we use the common setup of varying the number of slices (i.e., patients). In this scenario, we set \(K \in \{10, 20, 40, 80\}\). To ensure fair and accurate learning, either data‐selection or preprocessing strategies guided by the learned tensor components are essential for both \texttt{DDBN} and \texttt{CP-PAR}. Relying solely on data from patients with frequent visits would bias the model against those with fewer visits. Therefore, we truncate each patient’s dataset to the minimum number of visits across all patients, ensuring that sufficient information is captured consistently.

From Table~\ref{tab:causal_W}, we see that the SHD of the intra‑slice network recovered by our \texttt{CaRTeD} method is roughly half that of \texttt{DDBN} for all $K \neq 10$, indicating substantially more accurate structural recovery. For $K=10$, \texttt{DDBN} fails, marked as “NA”, which is unsurprising given the very limited patient and visit information in that case. Turning to the FDR, \texttt{CaRTeD} again outperforms \texttt{DDBN}, with an FDR roughly half that of \texttt{DDBN} in all cases. However, \texttt{DDBN}’s FDR increases as $K$ increases. In contrast, the FDR of \texttt{CaRTeD} remains stable around 0.24. Finally, the TPR of \texttt{CaRTeD} is three to four times higher than that of \texttt{DDBN}, confirming that \texttt{CaRTeD} yields more accurate recovery of the intra‑slice $W$. From Table~\ref{tab:causal_A}, we observe that \texttt{DDBN} fails to produce any inter-slice edges, indicating that separate learning does not capture the necessary temporal information. Although both methods yield similar SHD values, the TPR of \texttt{CaRTeD} is roughly three to four times higher than that of \texttt{DDBN} in all cases. As more patient data are included, the TPR of \texttt{CaRTeD} approaches 1. Finally, both methods exhibit relatively high FDRs—unsurprising given the difficulty of disambiguating time-crossing relations—but whereas the FDR of \texttt{CaRTeD} stabilizes around 0.7, that of \texttt{DDBN} rises more rapidly, resulting in faster performance degradation. We believe that the primary reason for \texttt{DDBN}’s poorer performance is its lower tensor decomposition accuracy, a consequence of the absence of joint regulation by causal-structure information. This underscores the importance of our joint-learning approach.

\section{Application}

In this section, we evaluate the performance of \texttt{CaRTeD} on a real-world dataset derived from the MIMIC-III electronic health record (EHR) \citep{Johnson2016MIMIC}, a publicly available and widely used resource in clinical research. This dataset contains detailed health information for over 40,000 ICU patients treated at the Beth Israel Deaconess Medical Center between 2001 and 2012, including demographics, medications, procedures, diagnoses, and mortality outcomes. For this study, we represent the EHR data as a third-order tensor with modes corresponding to hospital visits (mode-1), ICD-9 diagnosis codes (mode-2), and patients (mode-3). Each tensor entry $\mathcal{X}_{ijk}$ indicates the number of times a patient $k$ received diagnosis $j$ during visit $i$. Although this value is typically $0$ or $1$, occasionally it may show a value other than one during longer visits. By swapping the focus from diagnoses to medications or procedures, we can identify alternative phenotypes. To enhance interpretability, we preprocess the dataset by selecting only patients with at least three visits and retain the 202 most frequent ICD-9 codes among them, excluding codes beginning with 'V' or 'E' that denote supplementary information \citep{Kim2017Discriminative}. After preprocessing, the dataset consists of 2370 patients, 202 diagnostic features, and up to 42 hospital visits per patient. The resulting tensor has a non-zero element ratio of 0.0433. We apply both our method and the benchmark models to this processed data to extract medical phenotypes and the causal structure. For both benchmarks and \texttt{CaRTeD}, the hyperparameters of learning the causal structure are set as $\lambda_W = \lambda_A = 0.2$. To process the final causal graph, we set the a threshold of 0.03 for both $W, A$ (e.g., ignore the entries less than 0.03). In the extraction phase of the benchmark method, we apply only the non-negative constraint. Finally, we validate the results from both perspectives using either expert knowledge or authoritative medical literature.

\begin{table}[ht]
\centering
\caption{Phenotypes with the diagnoses (in ICD code form) extracted by \texttt{CaRTeD} and \texttt{COPA} from the MIMIC-III dataset.}
\label{tab:phenotypes}
\begin{tabular}{lccc}
\toprule
Phenotype  & CaRTeD & COPA \\
\midrule
\multirow{4}{*}{\textbf{Kidney disease}}
   & 5856 & 5856 \\
    & 40391 & 40391  \\
   & 28521 & 28521  \\
   & 3572  & 3572  \\ 
\midrule
\multirow{4}{*}{Hypertension \& hyperlipidemia}
   & 4019 & 4019   \\
   & 25000 & 25000  \\
    & 41401 & 2724  \\
   & 2724 & 41401  \\
\midrule

\multirow{4}{*}{Respiratory failure \& sepsis}
    & 5849 & 5849  \\
    & 99592 & 99592\\
    & 51881 & 51881  \\
    & 78552 & 78552   \\
\midrule
  
\multirow{4}{*}{Heart failure}
   & 4280 &  4280  \\
    & 42731 &  42731  \\
   & 41401 &  41401 \\
    & 40390 &  40390 \\

\bottomrule
\end{tabular}
\end{table}

\begin{figure}[H]
    \centering
    \includegraphics[width=0.4\linewidth]{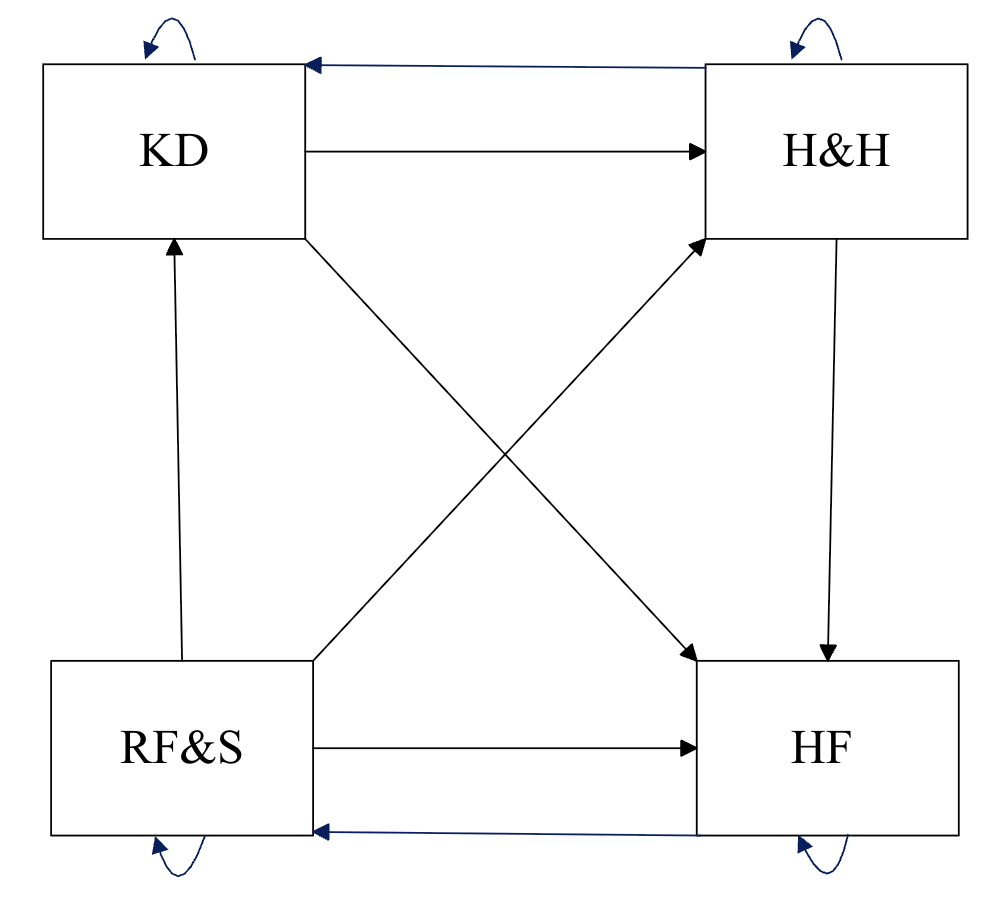}
    \caption{The summarized causal phenotype network generated by \texttt{CaRTeD}. KD denotes kidney disease; H\&H denotes hypertension and hyperlipidemia; HF denotes heart failure; and RF\&S denotes respiratory failure and sepsis.
}
    \label{CPNsum}
\end{figure}

We first show the phenotypes extracted by \texttt{CaRTeD} and \texttt{COPA}. We consider four phenotypes (i.e., \(R = 4\) and \(V \in \mathbb{R}^{202 \times 4}\)). To summarize each phenotype, we select the five largest values in each column of \(V\) and then choose the corresponding diagnoses for that phenotype. For example, for the phenotype defined as “kidney disease” in Table~\ref{tab:phenotypes}, we identify diagnoses such as end-stage renal disease (5856), hypertensive chronic kidney disease (40391), etc. Although our results illustrate four clusters, more than four meaningful phenotypes can be identified in practice. Lastly, a domain expert interprets the decomposed tensor and consolidates the results into clinically meaningful phenotypes. The extracted phenotypes are summarized in Table~\ref{tab:phenotypes} by selecting four diagnoses. In this table, we report the ICD-9 codes and describe each code in Table~\ref{tab:icd9_descriptions}. As illustrated in the table, our \texttt{CaRTeD} approach retrieves the same set of diagnostic codes per phenotype as \texttt{COPA}. This parity confirms that \texttt{CaRTeD} maintains the effectiveness of the underlying tensor decomposition.

\begin{table}[ht]
  \centering
  \caption{ICD-9 Codes and Descriptions}
  \label{tab:icd9_descriptions}
  \begin{tabular}{ll}
    \hline
    \textbf{Code (ICD-9)} & \textbf{Description} \\
    \hline
    4280   & Unspecified congestive heart failure \\
    42731  & Atrial fibrillation \\
    41401  & Coronary atherosclerosis of native coronary artery \\
    40390  & Unspecified hypertensive chronic kidney disease \\
    5849   & Acute kidney failure, unspecified \\
    40391  & Unspecified hypertensive chronic kidney disease \\
    4019   & Unspecified essential hypertension \\
    5859   & Unspecified chronic kidney diseased \\
    5990   & Unspecified urinary tract infection \\
    5856   & End-stage renal disease \\
    28521  & Anemia in chronic kidney disease \\
    3572   & Polyneuropathy in diabetes \\
    25000  & Diabetes mellitus without mention of complication \\
    2724   & Other and unspecified hyperlipidemia \\
    51881  & Acute respiratory failure \\
    99592  & Severe sepsis \\
    78552  & Septic shock \\
    \hline
  \end{tabular}
\end{table}

\begin{figure}[ht]
  \centering
  \begin{subfigure}[b]{0.45\textwidth}
    \centering
    \includegraphics[width=\textwidth]{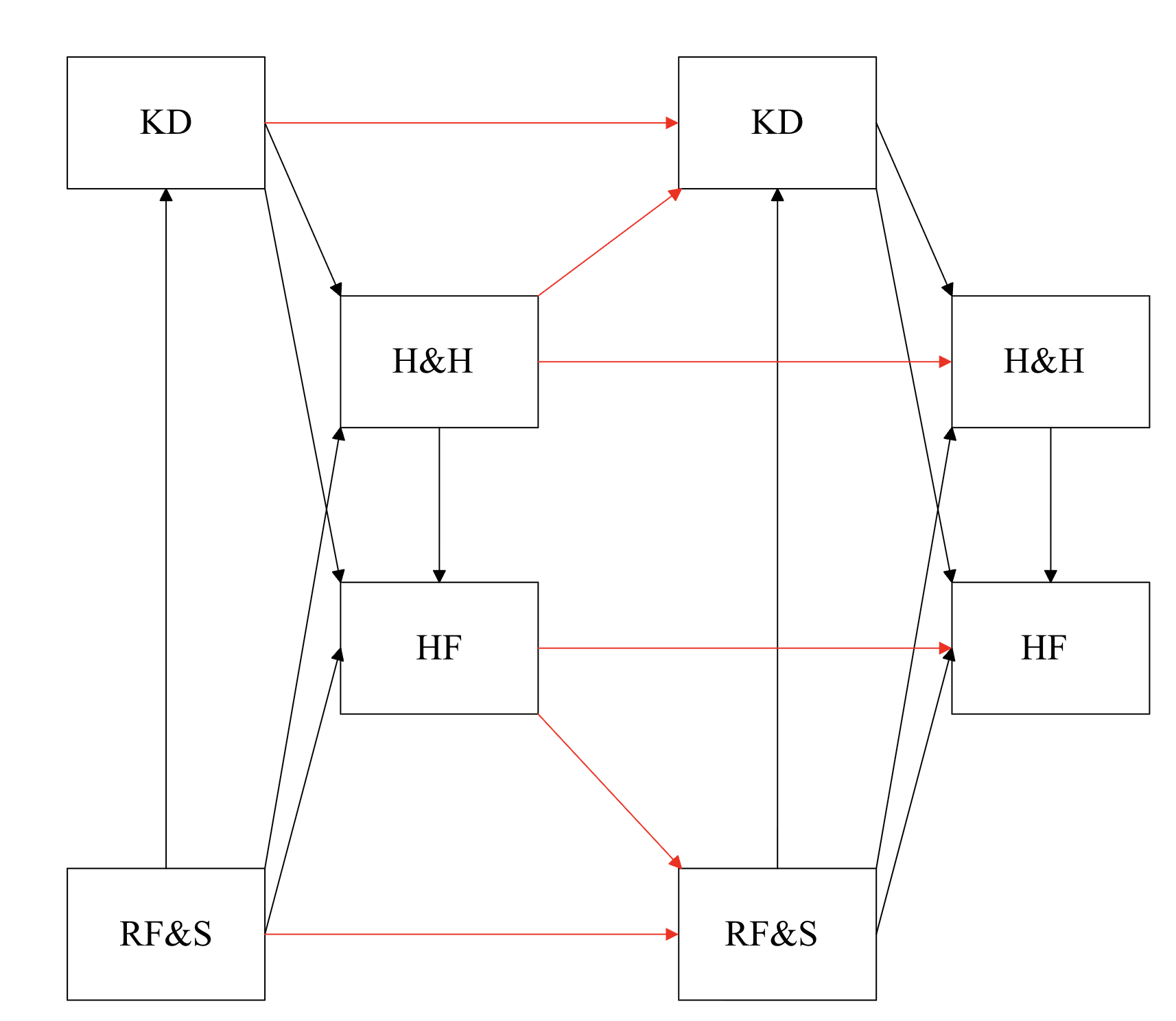}
    \caption{}
    \label{CPN:CaRTeD}
  \end{subfigure}
  \hfill
  \begin{subfigure}[b]{0.45\textwidth}
    \centering
    \includegraphics[width=\textwidth]{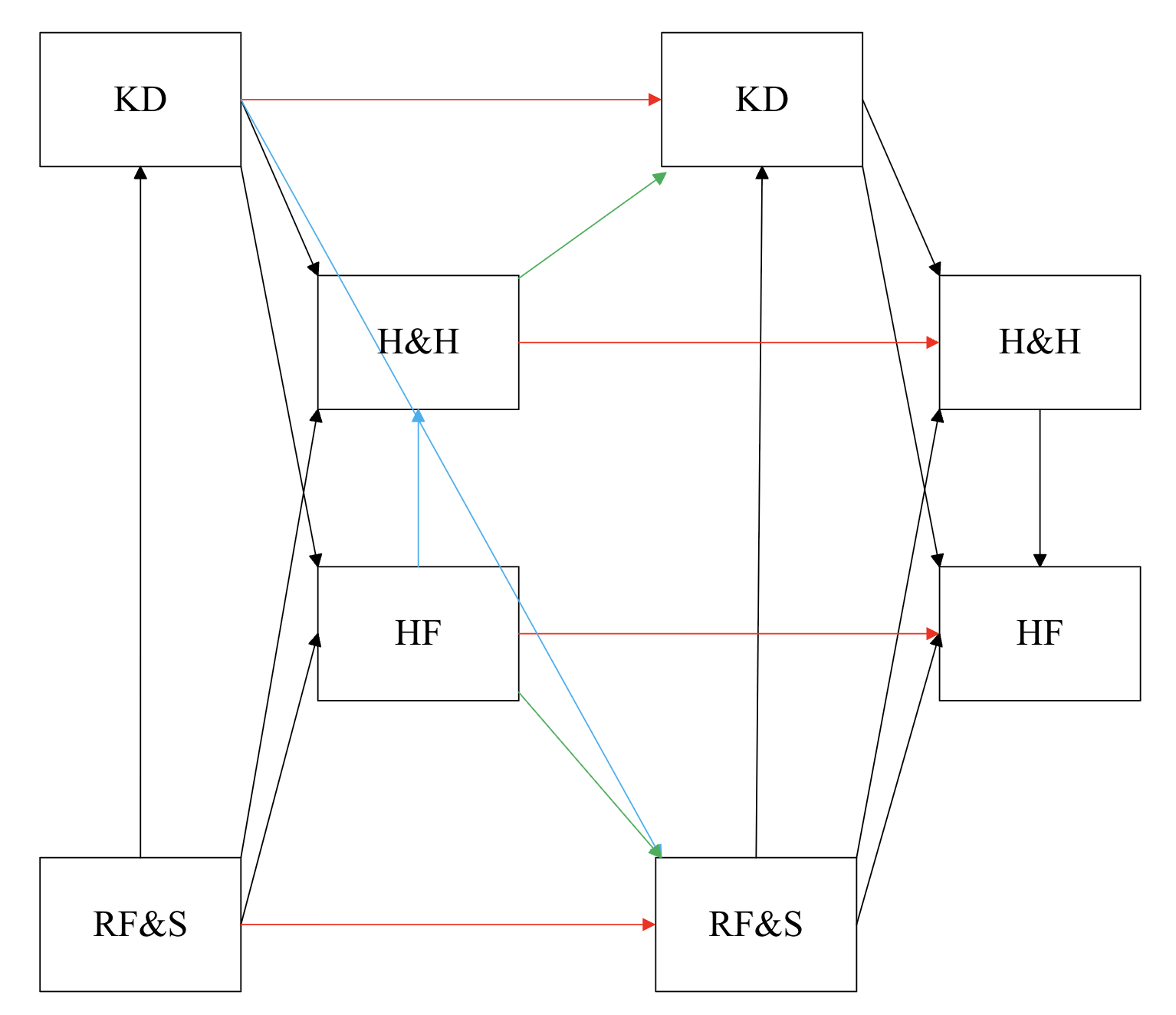}
    \caption{}
    \label{CPN:benckmark}
  \end{subfigure}
  \caption{$(a)$ is the inferred causal phenotype network by \texttt{CaRTeD}. $(b)$ is the inferred causal phenotype network by benchmark method. Red edges represent the inter slice and black edges represent the intra slice. The green edges and blue edges in $(b)$ represent the missing edge and the additional edges, compared with the $(a)$. KD denotes kidney disease; H\&H denotes hypertension and hyperlipidemia; HF denotes heart failure; and RF\&S denotes respiratory failure and sepsis.}
  \label{fig:both}
\end{figure}

More importantly, our method infers the causal network among those phenotypes simultaneously. An example of the resulting network is shown in Fig.~\ref{CPNsum}. To improve the readability, we assume that each node in the graph corresponds to a defined phenotype. Note that this is a summarized version of the temporal causal diagram, since the temporal stage only reveals the lesion or degradation rates (e.g., faster rates correspond to edges from \(W\)). \textbf{To the best of our knowledge, there is no ground-truth causal diagram among these phenotypes.} Therefore, it is difficult to directly validate our method against the benchmarks. Hence, we validate our results against evidence from the medical literature for each edge. To improve readability, we display the graph in two parts, one for inter-slice edges and the other for intra-slice edges, as shown in Fig.~\ref{fig:both}. Comparing the CPNs in Fig.~\ref{CPN:CaRTeD} and Fig.~\ref{CPN:benckmark}, we observe slight differences, two missing edges, one additional edge, and one reversed edge. Analyzing these edges further illustrates performance. In our paper, we adopt a high-specificity validation rule that retains only edges backed by strong clinical evidence and marks all others as errors. Additionally, we provide an example of post-processing to decide the final causal phenotype network, since a purely data-driven method yields only a Markov equivalence class. The inferred phenotype causal network by \texttt{CaRTeD} is shown in Fig.~\ref{CPN:CaRTeD}. We summarize the full CPN construction procedure in Supplementary Material \S\ref{CPN proc}.

To verify the results from \texttt{CaRTeD}, we first examine the inter‐slice causal diagram (highlighted by red edges). The graph shows that each phenotype follows its own temporal trajectory across visits, which is expected given our use of longitudinal EHR data. For example, a patient diagnosed with kidney disease at an early visit is likely to exhibit related symptoms in subsequent visits. Importantly, our inferred network captures clinically supported causal relationships. As reported by \citet{Burnier2023CR}, hypertension is a principal cause of chronic kidney disease. This is reflected in our causal graph (i.e., \textbf{Hypertension $\to$ Kidney disease} is in red). Similarly, \citet{IqbalGupta2023Cardiogenic} describe acute rises in left‐atrial pressure during decompensated heart failure force plasma ultrafiltrate into alveoli, producing cardiogenic pulmonary edema, a classic type I (hypoxemic) respiratory failure. Our network captures this pathophysiology via the edge \textbf{Heart failure $\to$ Respiratory failure}. In contrast, the benchmark method’s causal diagram fails to include these two key edges. Moreover, there is no evidence that respiratory failure causes chronic kidney disease, as discussed by \citet{YaxleyScott2019Respiratory}. Hence, \texttt{CaRTeD} provides a more accurate causal phenotype network.

Then, we verify the intra‐slice causal diagram (depicted by black edges) in Fig.~\ref{CPN:CaRTeD}. The graph indicates that kidney disease influences both hypertension and heart failure. This is consistent with clinical findings: for the edge \textbf{kidney $\to$ hypertension}, \citet{Siragy2010IntrarenalRAAS} show that chronic kidney disease (CKD) induces secondary hypertension via activation of the renin–angiotensin–aldosterone system (RAAS), sodium–water retention, and increased vascular resistance. Moreover, \citet{Segall2014HeartFailureCKD} report that heart failure (HF) is the leading cardiovascular complication in CKD patients, with prevalence rising as kidney function declines, supporting the edge \textbf{kidney $\to$ heart failure} in our graph. Clearly, our causal graph also correctly captures the relationship \textbf{hypertension $\to$ heart failure}, which has been supported in the medical literature (e.g., \citep{Nadar2021HeartInHypertension, Edward1992HIH}). In contrast, the benchmark method shown in Fig.~\ref{CPN:benckmark} reversed this direction, which is unreasonable. As explained by \citet{MartinPerez2019Development},  heart failure typically leads to hypotension (low blood pressure), not hypertension. Thus, the causal network produced by our \texttt{CaRTeD} is more accurate.

\begin{figure}[H]
  \centering
  \begin{subfigure}[b]{0.45\textwidth}
    \centering
    \includegraphics[width=\textwidth]{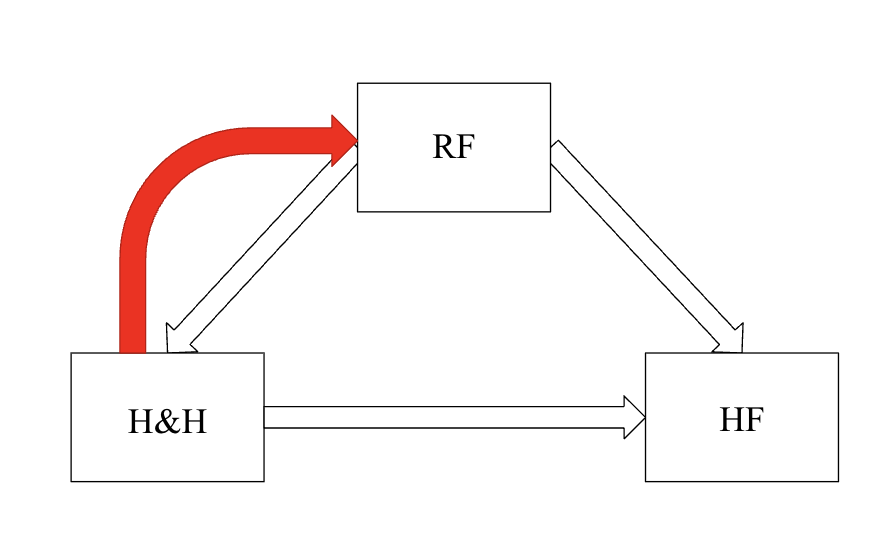}
    \caption{}
    \label{MEC:CaRTeD}
  \end{subfigure}
  \begin{subfigure}[b]{0.45\textwidth}
    \centering
    \includegraphics[width=\textwidth]{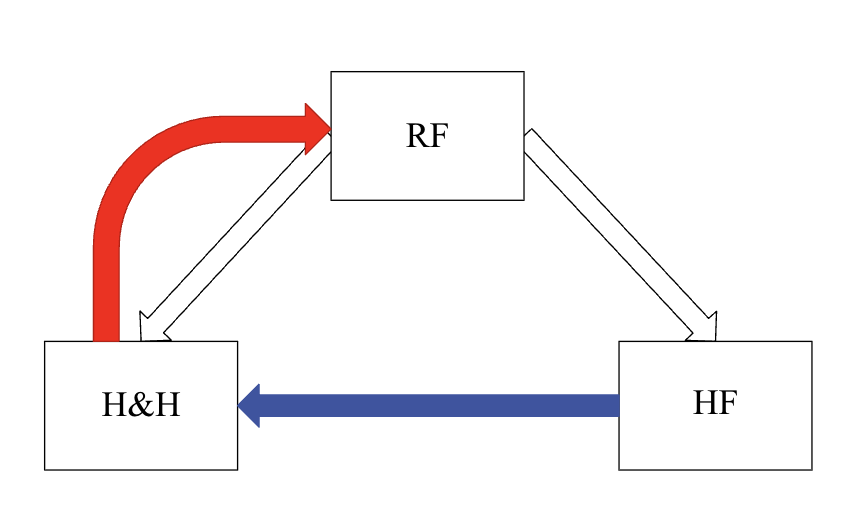}
    \caption{}
    \label{MEC:benckmark}
  \end{subfigure}
 \caption{Example of Markov equivalent class from our causal graph. The red arrow is corrected by the expert knowledge without making the cycle. H\&H denotes hypertension and hyperlipidemia; HF denotes heart failure; and RF denotes respiratory failure.}
  \label{fig:MEC_both}
\end{figure}

Finally, we find that respiratory failure and sepsis cause all other diseases. Studies by \citet{Kakihana2016SepsisMyocardial} and \citet{Antonucci2024Incidence} show that sepsis can lead to kidney and heart failure by shutting down the vital organs and inducing myocardial dysfunction. However, there is little evidence that it causes hypertension. We therefore consider the direction reversed. Note that our results represent a Markov equivalence class, a class of causal graphs that share the same conditional independencies. Accordingly, certain edges may be reversed without creating cycles, as illustrated in Fig.~\ref{MEC:CaRTeD}. To verify whether hypertension causes respiratory failure, case series in emergency medicine report that severe elevations in blood pressure can precipitate acute cardiogenic pulmonary edema, often termed sympathetic crashing acute pulmonary edema (SCAPE), a form of respiratory failure \citep{Sanjay2001APEA}. However, the benchmark method presents this edge in a reverse direction. As shown in Fig.~\ref{MEC:benckmark}, reversing the edge would introduce a cycle weakening the casual representation. Therefore, the CPN inferred by \texttt{CaRTeD} will be more interpretable.

\section{Discussion and Conclusion}

We propose \texttt{CaRTeD}, a joint‐learning framework for temporal causal structure and irregular tensor decomposition, and illustrate its application using electronic health record data (e.g., temporal causal phenotype network (tCPN) and computational phenotyping). Our framework addresses three key challenges. First, data from a single patient are insufficient to learn a meaningful tCPN. Second, unsupervised tensor decomposition methods lack dynamical or causal constraints. Third, directly applying causal representation learning without integrating the structure among meaningful latent clusters (e.g., phenotypes) yields limited insight. To overcome the last two challenges, we design an alternating optimization scheme that updates the tensor and causal blocks iteratively. To address the first challenge, we incorporate a state‐of‐the‐art aggregation approach across all slices (e.g., patients). More importantly, we present a theoretical analysis of our algorithm based on Lipschitz continuity, coercivity, first‐order optimality conditions, etc. In particular, we prove convergence for the optimization problem subject to the non-convex PARAFAC2 constraint. This analysis not only fills the gap in theoretical guarantees for the ADMM family applied to irregular tensor decomposition, but also provides additional insights and guidance for the design of related algorithms. Our experimental results demonstrate that the joint‐learning framework outperforms state‐of‐the‐art methods across six benchmark tests from two perspectives. In particular, under causally informed tensor decomposition, we demonstrate that our \texttt{CaRTeD} yields more accurate results. Furthermore, we show that a simple warm‐start initialization with a single component can deliver substantially greater computational efficiency and improved accuracy. Since we introduce a new problem in our paper, causal-informed tensor decomposition, we validate its feasibility and applicability through application to EHR-based phenotyping, a highly valuable data source in healthcare. Using our framework, we simultaneously learn computational phenotypes and their corresponding causal networks. The application results demonstrate that our method produces more accurate and explainable causal structures, facilitating straightforward post-processing. 

However, as an initial effort in this direction, our work has several limitations and opens up avenues for future investigation. Our framework presupposes a single, time-invariant Dynamic Bayesian Network structure, captured by the matrices $W$ and $A$, that holds for every time series in the data. Relaxing this restriction could be valuable, for example, by allowing the graph to evolve gradually over time \citep{LeTV2009}. Further research could also examine how the algorithm behaves with non-stationary or cointegrated series \citep{malinsky19a} or under hidden-confounder scenarios \citep{Huang2015ITD}. Moreover, as shown by \citet{chen2025fDBN}, learning causal structures for heterogeneous data (i.e., allowing for different underlying causal graphs) remains a highly promising research direction. In EHR datasets, it is plausible that distinct patient subgroups exhibit their own causal-phenotype networks. We notice that our choice of a linear model was made solely for clarity, which highlights the core dynamic and temporal features of the task. More expressive nonlinear relationships could be captured with Gaussian process models \citep{jin2019physics, shen2023multi, gnanasambandam2024deep, liu2024statistical} or neural‐network architectures. Likewise, the squared-error loss used in our method could be replaced with logistic (or, more generally, any exponential-family) likelihood to handle binary outcomes. Extending the framework to mixed continuous–discrete variables would also be valuable, as such data types are common in real-world settings \citep{Andrews2019LHd}.

\bibliographystyle{unsrtnat}
\bibliography{Reference}

\newpage
\appendix

\section{Proof}

\subsection{proof of Lemma.~\ref{lemma:Li-smooth}}
\begin{proof}

To prove that each $f^s_i$ is \(L_i\)-smooth, we need to show each of them has Lipschitz continuous gradient. Recall that the problem of $S_k$ is vectorized as follows:
$$
A=(V^{\!\top}\odot U_k)\in\mathbb R^{(I_kJ)\times R},
\qquad      
b=x_k\in\mathbb R^{I_kJ}   
$$

Then the vectorized problem can be written as    
\[
f(s)
=\tfrac12\|A s-b\|_{2}^{2}
\]

Since it is a quadratic term, we can have

\[\nabla f(s)=A^{\!\top}(A s-b) \quad \nabla^{2} f = A^{\!\top}A\]

For any $s,\hat s$,
$$
\|\nabla f(s)-\nabla f(\hat s)\|_2
  =\| (A^{\!\top}A)\,(s-\hat s)\|_2
  \le\|A^{\!\top}A\|_2\;\|s-\hat s\|_2.
$$

Thus, we can see that $f_i^s$ is $L_i$-smooth for all $i \in [K]$.
\end{proof}

\subsection{proof of Lemma.\ref{lemma:smooth_result}}

\begin{proof}
Recall that the standard augmented Lagrangian is given by the following. 
\begin{equation}
\mathcal L\bigl(\{S_k, \tilde S_k\},\mu_{k}\bigr)
=\sum_{k=1}^K f_k(S_k)+ h(\tilde S_k)
\;+\;\sum_{k=1}^K\langle \mu_{\tilde{S}_k},\,S_k - \tilde S_k\rangle
\;+\;\sum_{k=1}^K\frac{\rho_k}{2}\,\bigl\lVert S_k - \tilde S_k\bigr\rVert^2.
\end{equation}
When updating the $S_k$ block for $k \in [K]$. The first‐order optimality condition is
\[
\nabla f_k\bigl(S_k^{(t+1)}\bigr)+\mu^{(t)}_{\tilde{S}_k}\;+\;\rho_k\bigl(S_k^{(t+1)} - \tilde{S}_k^{(t+1)}\bigr)=0
\]
Combining this with the dual‐update step,

\begin{equation}
    \begin{aligned}   &\mu^{(t+1)}_{\tilde{S}_k}=\mu^{(t)}_{\tilde{S}_k}+\rho_k\bigl(S_k^{(t+1)}- \tilde S_k^{(t+1)} \bigr),\\
    \implies &\nabla f_k\bigl(S_k^{(t+1)}\bigr)=-\mu^{(t+1)}_{\tilde{S}_k}
    \end{aligned}
\end{equation}
By Lemma.\ref{lemma:Li-smooth}, \(f^s_k\) is \(L_k\)-smooth. Therefore, we can observe the desired result 
\[
\bigl\lVert \mu_{\tilde{S}_k}^{(t+1)}-\mu_{\tilde{S}_k}^{(t)}\bigr\rVert
=\bigl\lVert \nabla f_k(S_k^{(t+1)})-\nabla f_k(S_k^{(t)})\bigr\rVert
\;\le\;L_k\,\bigl\lVert S_k^{(t+1)}-S_k^{(t)} \bigr\rVert
\]
\end{proof}

\subsection{proof of Lemma.\ref{lemma:S_ bound}}

\begin{proof}
We first split the successive difference of the augmented Lagrangian by
\begin{align*}
&\mathcal L\bigl(\{S_k^{(t+1)},\,\tilde S_k^{(t+1)}\},\,\mu^{(t+1)}_{\tilde S_k}\bigr)
-\mathcal L\bigl(\{S_k^{(t)},\,\tilde S_k^{(t)}\},\,\mu^{(t)}_{\tilde S_k}\bigr)\\
&\quad=
\Bigl[
\mathcal L\bigl(\{S_k^{(t+1)},\,\tilde S_k^{(t+1)}\},\,\mu^{(t+1)}_{\tilde S_k}\bigr)
-\,\mathcal L\bigl(\{S_k^{(t+1)},\,\tilde S_k^{(t+1)}\},\,\mu^{(t)}_{\tilde S_k}\bigr)
\Bigr]\\
&\qquad+
\Bigl[
\mathcal L\bigl(\{S_k^{(t+1)},\,\tilde S_k^{(t+1)}\},\,\mu^{(t)}_{\tilde S_k}\bigr)
-\mathcal L\bigl(\{S_k^{(t)},\,\tilde S_k^{(t)}\},\,\mu^{(t)}_{\tilde S_k}\bigr)
\Bigr].
\end{align*}

To improve the readability, we write $\mu_{\tilde S_k} = \mu_{k}$. The bound for the first term is 
\begin{align*}
\mathcal L\bigl(&\{S_k^{(t+1)},\,\tilde S_k^{(t+1)}\},\,\mu_k^{(t+1)}\bigr)
-\mathcal L\bigl(\{S_k^{(t+1)},\,\tilde S_k^{(t+1)}\},\,\mu_k^{(t)}\bigr)\\
&=\sum_{k=1}^K
\bigl\langle
\mu_k^{(t+1)}, S_k^{(t+1)} - \tilde S_k^{(t+1)} 
\bigr\rangle - \sum_{k=1}^K
\bigl\langle
\mu_k^{(t)}, S_k^{(t+1)} - \tilde S_k^{(t+1)} 
\bigr\rangle\\
&=\sum_{k=1}^K
\bigl\langle
\mu_k^{(t+1)} - \mu_k^{(t)}, S_k^{(t+1)} - \tilde S_k^{(t+1)} 
\bigr\rangle\\
&= \sum_{k = 1}^K
\frac{1}{\rho_k}
\bigl\lVert
\mu_k^{(t+1)} - \mu_k^{(t)}
\bigr\rVert^2
\end{align*}

To show the bound for the second term, we can split it again as: 

\begin{equation}
\label{lm3:eqst}
\begin{aligned}
&\mathcal L\bigl(\{S_k^{(t+1)},\,\tilde S_k^{(t+1)}\},\,\mu_k^{(t)}\bigr)
-\mathcal L\bigl(\{S_k^{(t)},\,\tilde S_k^{(t)}\},\,\mu_k^{(t)}\bigr)\\
&\quad=
\Bigl[
\mathcal L\bigl(\{S_k^{(t+1)},\,\tilde S_k^{(t+1)}\},\,\mu_k^{(t)}\bigr)
-\mathcal L\bigl(\{S_k^{(t)},\,\tilde S_k^{(t+1)}\},\,\mu_k^{(t)}\bigr)
\Bigr]\\
&\qquad\;+\;
\Bigl[
\mathcal L\bigl(\{S_k^{(t)},\,\tilde S_k^{(t+1)}\},\,\mu_k^{(t)}\bigr)
-\mathcal L\bigl(\{S_k^{(t)},\,\tilde S_k^{(t)}\},\,\mu_k^{(t)}\bigr)
\Bigr]. 
\end{aligned}
\end{equation}

By strong convexity, the first term of Eq.\eqref{lm3:eqst} can be bounded as follows

\begin{equation}
\begin{aligned}
\mathcal L\bigl(\{S_k^{(t)},\,\tilde S_k^{(t+1)}\},\,\mu_k^{(t)}\bigr)
 & \geq \mathcal L\bigl(\{S_k^{(t+1)},\,\tilde S_k^{(t+1)}\},\,\mu_k^{(t)}\bigr) \\& + \sum_{k=1}^{K} \langle \nabla_{S_k}\mathcal L\bigl(\{S_k^{(t+1)},\,\tilde S_k^{(t+1)}\},\,\mu_k^{(t)}\bigr),S_k^{(t)} - S_k^{(t+1)} \rangle \\&
 +\frac{\gamma_k(\rho_k)}{2}\,\bigl\lVert S_k^{(t)} - S_k^{(t+1)} \bigr\rVert^2
\end{aligned}
\end{equation}

Hence, we can have 

\begin{align*}
&\mathcal L\bigl(\{S_k^{(t+1)},\tilde S_k^{(t+1)}\},\mu_k^{(t)}\bigr)
-
\mathcal L\bigl(\{S_k^{(t)},\tilde S_k^{(t+1)}\},\mu_k^{(t)}\bigr) \\
&\leq
\sum_{k=1}^{K}
\Bigl\langle\nabla_{S_k}\mathcal L,\,S_k^{(t+1)}-S_k^{(t)}\Bigr\rangle
-\frac{ \gamma_k(\rho_k)}{2}\,
      \|S_k^{(t+1)}-S_k^{(t)}\|^{2}.
\end{align*}

Similarly, for the second term of Eq.\eqref{lm3:eqst}, we can have the following: 
\begin{align*}
&\mathcal L\bigl(\{S_k^{(t)},\tilde S_k^{(t+1)}\},\mu_k^{(t)}\bigr)
\;-\mathcal L\bigl(\{S_k^{(t)},\tilde S_k^{(t)}\},\mu_k^{(t)}\bigr)
\\&\leq\; \sum_{k=1}^{K}\bigl\langle
      \nabla_{\tilde S_k}\mathcal L,
      \,\tilde S_k^{(t+1)}-\tilde S_k^{(t)}
  \bigr\rangle
\;-\;
\frac{\tilde \gamma_k(\rho_k)}{2}\,
\bigl\lVert
      \tilde S_k^{(t+1)}-\tilde S_k^{(t)}
\bigr\rVert
\end{align*}

Thus, we can see that 
\begin{align*}
    \mathcal L\bigl(\{S_k^{(t+1)},\,\tilde S_k^{(t+1)}\},\,\mu_k^{(t)}\bigr)
& -\mathcal L\bigl(\{S_k^{(t)},\,\tilde S_k^{(t)}\},\,\mu_k^{(t)}\bigr)\\ 
& \leq \sum_{k=1}^{K}
\Bigl\langle\nabla_{S_k}\mathcal L,\,S_k^{(t+1)}-S_k^{(t)}\Bigr\rangle
-\frac{\gamma_k(\rho_k)}{2}\, \|S_k^{(t+1)}-S_k^{(t)}\|^{2}.\\
&\qquad +\bigl\langle
      \nabla_{\tilde S_k}\mathcal L,
      \,\tilde S_k^{(t+1)}-\tilde S_k^{(t)}
  \bigr\rangle
\;-\;
\frac{\tilde \gamma_k(\rho_k)}{2}\,
\bigl\lVert
      \tilde S_k^{(t+1)}-\tilde S_k^{(t)}
\bigr\rVert^{2},\\
& \leq  \sum_{k=1}^{K}
-\frac{\gamma_k(\rho_k)}{2}\, \|S_k^{(t+1)}-S_k^{(t)}\|^{2}- 
\frac{\tilde \gamma_k(\rho_k)}{2}\,
\bigl\lVert
      \tilde S_k^{(t+1)}-\tilde S_k^{(t)}
\bigr\rVert^{2},
\end{align*}
The last inequality is hold since we have used the optimality of each subproblem that satisfies the optimality condition. By Lemma.\ref{lemma:smooth_result}, we can combine each term as 
\begin{align*}
&\mathcal L\bigl(\{S_k^{(t+1)}, \tilde S_k^{(t+1)}\},\mu^{(t+1)}_{\tilde{S}_k}\bigr) 
- 
\mathcal L\bigl(\{S_k^{(t)}, \tilde S_k^{(t)}\},\mu^{(t)}_{\tilde{S}_k}\bigr)\\
&\leq \sum_{k=1}^{K}
-\frac{\gamma_k(\rho_k)}{2}\, \|S_k^{(t+1)}-S_k^{(t)}\|^{2}- 
\frac{\tilde \gamma_k(\rho_k)}{2}\,
\bigl\lVert
      \tilde S_k^{(t+1)}-\tilde S_k^{(t)}
\bigr\rVert_F^{2} + \frac{1}{\rho_k}
\bigl\lVert
\mu_k^{(t+1)} - \mu_k^{(t)}
\bigr\rVert^2\\
&\leq \sum_{k=1}^{K}
(\frac{L_k^2}{\rho_k} -\frac{\gamma_k(\rho_k)}{2}) \|S_k^{(t+1)}-S_k^{(t)}\|^{2}- 
\frac{\tilde \gamma_k(\rho_k)}{2}\,
\bigl\lVert \tilde S_k^{(t+1)}-\tilde S_k^{(t)}
\bigr\rVert^{2}
\end{align*}
\end{proof}

\subsection{proof of Theorem.\ref{Thm:convegent}}
\begin{proof}
Recall the Lagrangian form, we have 
\begin{equation}
\begin{aligned}
&\mathcal L\bigl(\{S_k^{(t+1)},\,\tilde S_k^{(t+1)}\},\,\mu_{\tilde S_k}^{(t+1)}\bigr)\\
&=\sum_{k = 0}^K h(\tilde S_k^{(t+1)})+f_k\bigl(S_k^{(t+1)}\bigr)
\;+\;\Bigl\langle \mu_{\tilde S_k}^{(t+1)},\,S_k^{(t+1)}-\tilde S_k^{(t+1)}\Bigr\rangle
\;+\;\frac{\rho_k}{2}\bigl\lVert S_k^{(t+1)}-\tilde S_k^{(t+1)}\bigr\rVert^2\\
& = \sum_{k = 0}^K
h(\tilde S_k^{(t+1)})+
f_k\bigl(S_k^{(t+1)}\bigr)
\;+\;\Bigl\langle \nabla f_k\bigl(\tilde S_k^{(t+1)}\bigr),\,\tilde S_k^{(t+1)} - S_k^{(t+1)}\Bigr\rangle
\;+\;\frac{\rho_k}{2}\bigl\lVert S_k^{(t+1)}-\tilde S_k^{(t+1)}\bigr\rVert^2
\end{aligned}
\end{equation}

The second equality hold since we can observe that $\nabla f_k\bigl(S_k^{(t+1)}\bigr)=-\mu^{(t+1)}_{\tilde{S}_k}$ and the outer product property. By Lemma.\ref{lemma:Li-smooth}, we can the following inequality. 

\begin{equation}
\begin{aligned}
f_k\bigl(S_k^{(t+1)}\bigr)
\;+\;\Bigl\langle \nabla f_k\bigl(\tilde S_k^{(t+1)}\bigr),\,\tilde S_k^{(t+1)} - S_k^{(t+1)}\Bigr\rangle
\;+\;\frac{\rho_k}{2}\bigl\lVert S_k^{(t+1)}-\tilde S_k^{(t+1)}\bigr\rVert^2
\ge
f_k\bigl(\tilde S_k^{(t+1)}\bigr)
\end{aligned}
\end{equation}
Therefore, we can have the following
\begin{equation}
    \begin{aligned}
    &\mathcal L\bigl(\{S_k^{(t+1)},\,\tilde S_k^{(t+1)}\},\,\mu_{\tilde S_k}^{(t+1)}\bigr) \geq \sum_{k = 0}^K
    h(\tilde S_k^{(t+1)})+f_k\bigl(\tilde S_k^{(t+1)}\bigr)
    \;=\;
    g\bigl(\tilde S_k^{(t+1)}\bigr)\,,   
    \end{aligned}
\end{equation}
Since the $h, f_k$ are all the function of Frobenius norm in our algorithm, we can know it is bounded below. therefore, the $\mathcal L\bigl(\{S_k^{(t+1)},\,\tilde S_k^{(t+1)}\},\,\mu_{\tilde S_k}^{(t+1)}\bigr)$ is bounded below as well. By Lemma.\ref{lemma:S_ bound}, we can say that $\mathcal L\bigl(\{S_k^{(t+1)},\,\tilde S_k^{(t+1)}\},\,\mu_{\tilde S_k}^{(t+1)}\bigr)$ is monotonically decreasing and convergent when the penalty parameters are chosen large enough. 
\end{proof}

\subsection{proof of Lemma.\ref{lemma:Uk bounded}}

\begin{proof}
    For the first two statements, the proofs can be trivially followed by Lemma.\ref{lemma:S_ bound} and Theorem.\ref{Thm:convegent}. For the last statement, we firstly discuss about the boundedness of $\hat U_k$. Since $\hat U_k = Q_kH$, we know that $Q_k$ is compact because of the orthonormality and $H$ is a fixed matrix during the updates of all other variables, therefore, the $\hat U_k$ is generated by the continuous mapping of the compact set, which is compact. However, the $H$ will be updated iteratively. The boundedness of $H^{(t)}$ should be verified. By our algorithm, for sufficient large $\rho_k$, the subproblem of $f_k^u(U_k)$ is strongly convex with modulus. Therefore,  by the Theorem.\ref{Thm:convegent} and Lemma.\ref{lemma:Li-smooth}, we can observe that 
    \[
    \| U_k^{(t+1)} - U_k^{(t)}\| \to 0 \quad \|  \mu_{\hat U_k}^{(t+1)} - \mu_{\hat U_k}^{(t)}\|\to 0
    \]
    Since the descent inequality by Lemma.\ref{lemma:S_ bound} This implies that 
    \[
    \sum_{t=0}^\infty\| U_k^{(t+1)} - U_k^{(t)}\| < \sum_{t=0}^\infty \mathcal L^{(t)} - \mathcal L^{(t+1)} = \mathcal L^0 - \mathcal L^* <\infty 
    \]
    Then, by Lemma.\ref{lemma:smooth_result}, the following holds for $\mu_{\hat U_k}$
    \[
    \sum_{t=0}^\infty\|\mu_{\hat U_k}^{(t+1)} - \mu_{\hat U_k}^{(t)}\| \leq \sum_{t=0}^\infty L_k \| U_k^{(t+1)} - U_k^{(t)}\| < \infty
    \]
    For our updating rule for $H^{(t+1)}$, we can have the following:
    \begin{equation}
    \begin{aligned}
     &\|H^{(t+1)} - H^{(t)}\|
    \;\le\;
    \sum_{k=1}^{K}\Bigl(
       \|U_k^{(t+1)} - U_k^{(t)}\|
     + \|\mu_{\hat U_k}^{(t)} - \mu_{\hat U_k}^{(t-1)}\|
    \Bigr)\\
    \implies & 
   \sum_{t=0}^\infty\bigl\|H^{(t+1)}-H^{(t)}\bigr\|
   \;\le\;
   \sum_{k=1}^K \Bigl(
     \sum_{t=0}^\infty\bigl\|U_k^{(t+1)}-U_k^{(t)}\bigr\|
     +\sum_{t=0}^\infty\bigl\|\mu_{\hat U_k}^{(t)}-\mu_{\hat U_k}^{(t-1)}\bigr\|
   \Bigr)
   \;<\;\infty.   
    \end{aligned}  
    \end{equation}
    Therefore, we can see that $H^{(t)}$ is a Cauchy sequence because of a finite telescoping sum. This implies that $H$ is bounded and 
    $S$ is bounded over $t$. Then, for boundedness of $U$, by 1 and 2, we know that $\mathcal{L}((U^{t}_k,\,\hat{U}^{t}_k,\,\mu^{t}_{\hat{U}_k})$ is upper bounded by the initial points, $(U^{0}_k,\,\hat{U}^{0}_k,\,\mu^{0}_{\hat{U}_k})$. 
    Therefore, $f_k^u(U_k)$ is bounded by $\mathcal{L}(U^{t}_k,\,\hat{U}^{t}_k,\,\mu^{t}_{\hat{U}_k})$. Note that the each $\hat U^{(t)}_k$ lies in a bounded set and the $f_k^u(U_k)$ is bounded below on that set. By the Defn.\ref{coercivity}, 
    $U_k$ is bounded. The boundedness of $\mu_{\hat U_k}$ can be derived by the optimality condition, which is used to prove lemma.\ref{lemma:smooth_result}. 
    \[
    \mu^{(t)}_{\hat{U}_k} = -\nabla f_k\bigl(U_k^{(t)}\bigr)
    \]
\end{proof}

\subsection{proof of Lemma.\ref{Lemma: Uk Subgradient bound}}
\begin{proof}
Given the following function
\begin{equation}
\mathcal L\bigl(\{U_k, \hat U_k\},\mu_{\hat{U}_k}\bigr)
=\sum_{k=1}^K f_k^u(U_k) + \iota_S (\hat U_k)+ \sum_{k=1}^K\langle \mu_{\hat{U}_k},\,U_k - \hat U_k\rangle
\;+\;\frac{\rho_k}{2}\,\bigl\lVert U_k - \hat U_k\bigr\rVert^2.
\end{equation}
we know 
\[
\partial \mathcal{L}\bigl(U_k^{t+1},\,\hat U_k^{t+1},\,\mu_{\hat U_k}^{t+1}\bigr)
=
\Biggl( \nabla_{ U_k}\mathcal{L},\;
  \nabla_{\hat U_k}\mathcal{L},\;
  \nabla_{\mu_{\hat U_k}}\mathcal{L}
\Biggr)
\bigl(U_k^{t+1},\,\hat U_k^{t+1},\,\mu_{\hat U_k}^{t+1}\bigr)\,.
\]
To prove this lemma, we need to show that each block of $\partial\mathcal{L}$ can be controlled by some constant depending on $\rho$. For $\mu_{\hat U_k}$ block, we have 
\[
\nabla_{\mu_{\hat U_k}}\mathcal{L} = \sum_kU_k^{(t+1)} - \hat U_k^{(t+1)} = \sum_k\frac{1}{\rho}(\mu^{(t+1)}_{\hat U_k} - \mu^{(t)}_{\hat U_k}). 
\]
By Lemma.\ref{lemma:smooth_result}, we have $\|\nabla_{\mu_{\hat U_k}}\mathcal{L}\| \leq \sum_k \frac{L_k}{\rho} \| U_k^{(t+1)} - U_k^{(t)}\|$. Then, for the $U_k$ block, we have the gradient 
\[
\nabla_{U_k}\mathcal{L} = \nabla f_k^u(U_k^{(t+1)}) + \mu^{(t+1)}_{\hat{U}_k} + \rho (U_k^{(t+1)} - \hat U_k^{(t+1)})
\]
Note that $\rho (U_k^{(t+1)} - \hat U_k^{(t+1)}) = \mu^{(t+1)}_{\hat U_k} - \mu^{(t)}_{\hat U_k}$ and $ \nabla f_k^u(U_k^{(t+1)}) = -\mu^{(t+1)}_{\tilde{U}_k}$. Following Lemma.\ref{lemma:smooth_result}, We can have that $\|\nabla_{U_k}\mathcal{L}\| = \|\mu^{(t+1)}_{\hat U_k} - \mu^{(t)}_{\hat U_k} \| \leq L_k \,\bigl\lVert U_k^{(t+1)} - U_k^{(t)}\bigr\rVert$. Finally, for the $\hat U_k$ and for all $s = \{1, 2,\ldots, K\}$, we observe the following
\begin{equation}
    \begin{aligned}
    & \frac{\partial\mathcal{L}}{\partial\hat U_k}  (\{U_k^{(t+1)}, \hat U_k^{(t+1)}\},\mu^{(t+1)}_{\hat{U}_k})  \\
    & = \partial_s \iota_S (\hat U^{(t+1)}_s) + \mu^{(t+1)}_{\hat U_k} + \rho (U_s^{(t+1)} - \hat U_s^{(t+1)}) \\
    & = \partial_s \iota_S (\hat U^{(t+1)}_s) + \mu^{(t)}_{\hat U_k} + \rho (U^{(t)}_{s}-\hat U^{(t+1)}_{\leq s} - \hat U^{(t)}_{> s})\\
    &+ \mu^{(t+1)}_{\hat U_k} - \mu^{(t)}_{\hat U_k} + \rho(-\hat U^{(t+1)}_{>s} + \hat U^{(t)}_{> s} -U^{(t)}_{s} +U^{(t+1)}_{s})
    \end{aligned}
\end{equation}
By the first order optimal condition on $\hat U^{(t+1)}_k$, we can have $0 \in \partial_s \iota_S (\hat U^{(t+1)}_s) + \mu^{(t)}_{\hat U_k} + \rho (U^{(t)}_{s}-\hat U^{(t+1)}_{\leq s} - \hat U^{(t)}_{> s})$.
Thus, we can have 
\[d_s = \mu^{(t+1)}_{\hat U_k} - \mu^{(t)}_{\hat U_k} + \rho(-\hat U^{(t+1)}_{>s} + \hat U^{(t)}_{> s} -U^{(t)}_{s} +U^{(t+1)}_{s}) \in \frac{\partial \mathcal{L}}{\partial\hat U_k}  (\{U_k^{(t+1)}, \hat U_k^{(t+1)}\},\mu^{(t+1)}_{\hat{U}_k})  \]
Therefore, we can have 
\begin{equation}
    \begin{aligned}
        \|d_s\| &\leq L_k \,\bigl\lVert U_k^{(t+1)} - U_k^{(t)}\bigr\rVert+ \rho (\sum_{k} \| \hat U^{(t+1)}_{k} - \hat U^{(t)}_{k}\| + \| U^{(t+1)}_{k} -U^{(t)}_{k}  \| )\\
        &\leq (L_k+ \rho) (\sum_{k} \| \hat U^{(t+1)}_{k} - \hat U^{(t)}_{k}\| + \| U^{(t+1)}_{k} -U^{(t)}_{k} \|)
    \end{aligned}
\end{equation}
Therefore, we have proved the statement.
\end{proof}

\subsection{proof of Lemma.\ref{lemma: Uk limit cont}}

\begin{proof}
By Lemma.\ref{lemma:S_ bound} and Theorem.\ref{Thm:convegent}, we can conclude that the $\mathcal{L}(U_k^{t_s},\hat U_k^{t_s},\mu_{\hat U_k}^{t_s})$ is monotonic decreasing and lower bounded, which implies the convergence. $\mathcal{L}$ is lower‐semicontinuous since it contains a indicator function, which is lower semicontinuous for a closed set. By the fact that the indicator function has discontinuous terms, we can have 
\begin{equation}
    \begin{aligned}
    &\lim_{s\to\infty}\mathcal{L}(U_k^{t_s},\hat U_k^{t_s},\mu_{\hat U_k}^{t_s})  \geq\mathcal{L}(U_k^*,\hat U_k^*,\mu_{\hat U_k}^*)
    \\  
     \implies &\lim_{s\to\infty}\mathcal{L}(U_k^{t_s},\hat U_k^{t_s},\mu_{\hat U_k}^{t_s})  - \mathcal{L}(U_k^*,\hat U_k^*,\mu_{\hat U_k}^*) \leq \lim_{ s\to\infty}\sup \iota_S(\hat U_k^{t_s}) - \iota_S(\hat U_k^*)     
    \end{aligned}
\end{equation}
Given that $ \hat U^{t_s}_k$ is the optimal solution for the sub-problem 

\[
\min_{\hat U^{t_s}_i} \mathcal{L}(U^{t_s-1},\hat U_{< k}^{t_s},\hat U_{k}^{t_s},\hat U_{> k}^{t_s-1},\mu_{\hat U_k}^{t_s-1}).
\]
Therefore, for any candidate (in particular \(\hat U_k^*\)) we have

\[
\mathcal{L}\!\bigl(U_k^{t_s-1},\widehat U_{<k}^{\,t_s-1},
                  \widehat U_k^{\,t_s},
                  \widehat U_{>k}^{\,t_s-1},
                  \mu_{\widehat U_k}^{\,t_s-1}\bigr)
\ \le\
\mathcal{L}\!\bigl(U_k^{t_s-1},\widehat U_{<k}^{\,t_s-1},
                  \widehat U_k^*,
                  \widehat U_{>k}^{\,t_s-1},
                  \mu_{\widehat U_k}^{\,t_s-1}\bigr).
\]
By taking the limits over the different between them, we can have $\limsup_{ s\to\infty}\iota_S(\hat U_k^{t_s}) - \iota_S(\hat U_k^*) \leq 0$. Therefore, the claim is proved.
\end{proof}

\subsection{proof of Theorem.\ref{Thm: Uk_con}}
\begin{proof}
To prove this statement, we only need to prove  $0 \in \partial \mathcal{L}((U^{*}_k,\,\hat{U}^{*}_k,\,\mu^{*}_{\hat{U}_k})$, which is standard \citep{Yu2019ADMM, Xu2013BCD, Hedy2011CDM}. By Lemma.\ref{lemma:Uk bounded}, we have shown that $(U^{t}_k,\,\hat{U}^{t}_k,\,\mu^{t}_{\hat{U}_k})$ is bounded, so there exist a convergent subsequence and a limit point such that 
\[
\lim_{s \to \infty}  (U_k^{t_s},\hat U_k^{t_s},\mu_{\hat U_k}^{t_s}) =  (U_k^*,\hat U_k^*,\mu_{\hat U_k}^*) 
\]
Then, by Lemma.\ref{lemma:Uk bounded} and Lemma.\ref{lemma:S_ bound}, the $\mathcal L(U^{t}_k,\,\hat{U}^{t}_k,\,\mu^{t}_{\hat{U}_k})$ is monotonically
decreasing and lower bounded. Therefore, 
\[
\lim_{t \to \infty}\|U_k^{(t)} -U_k^{(t+1)}\| = 0, \quad \text{and} \quad\lim_{t \to \infty}\|\hat U_k^{(t)} -\hat U_k^{(t+1)}\| = 0.
\]
From Lemma.\ref{Lemma: Uk Subgradient bound}, we have that there exists $d^k \in \partial\mathcal{L}(U^{k},\hat U^{k},\mu_{\hat{U}_k}^{k})$ such that $\|d^k\| \to 0$. Hence,  
\[
\lim_{s\to \infty} \| d^{k_s} \| = 0 
\]
Finally, by Lemma.\ref{lemma: Uk limit cont}, we have that 
\[
\lim_{s\to\infty}\mathcal{L}(U_k^{t_s},\hat U_k^{t_s},\mu_{\hat U_k}^{t_s}) =  \mathcal{L}(U_k^*,\hat U_k^*,\mu_{\hat U_k}^*) 
\]
By the sub-gradient definition \citep{Rockafellar1998VA}, we can have that $0 \in \partial \mathcal{L}(U_k^*,\hat U_k^*,\mu_{\hat U_k}^*)$
\end{proof}

\section{Closed form Procedure}
\subsection{Closed form of $U$}
\label{Up_U}
\[
\min_{U_k}\ 
\frac12\bigl\lVert X_k - U_k S_k V^\top\bigr\rVert_F^2
\;+\;
\frac{\rho_k}{2}\bigl\lVert U_k - C_1\bigr\rVert_F^2
+\frac{\rho_k}{2}\bigl\lVert U_k - C_2\bigr\rVert_F^2,
\]
where
\[
C_1 = \tilde U_k^{(t)} - \mu_{\tilde U_k}^{(t)},
\qquad
C_2 = \hat U_k^{(t)} - \mu_{\hat U_k}^{(t)}.
\]
Taking the derivative with respect to \(U_k\) gives
\begin{align}
&-2\,(X_k - U_k S_k V^\top)\,V S_k^\top
\;+\;\rho_k\,(U_k - C_1)\;+\;\rho_k\,(U_k - C_2)
\;=\;0.
\end{align}

\[
2\,U_k\,(S_k V^\top V S_k^\top) \;+\;2\,\rho_k\,U_k
\;=\;
2\,X_k\,V\,S_k^\top \;+\;\rho_k\,(C_1 + C_2).
\]

\[
U_k \bigl(S_k V^\top V S_k^\top + \rho_k I\bigr)
\;=\;
X_k V S_k^\top
\;+\;
\frac{\rho_k}{2}\,(C_1 + C_2).
\]
Finally, the closed‐form update is
\[
\boxed{
U_k 
=
\Bigl(
X_k V S_k^\top 
\;+\;
\tfrac{\rho_k}{2}\bigl(\tilde U_k^{(t)} + \hat U_k^{(t)} 
                   \;-\;\mu_{\tilde U_k}^{(t)} - \mu_{\hat U_k}^{(t)}\bigr)
\Bigr)
\bigl(S_k V^\top V S_k^\top + \rho_k I\bigr)^{-1}.
}
\]
\subsection{Closed form of $\tilde{U}_k$}
\label{Up_tU_k}

Since the given problem is:  
\[
\min_{\tilde{U}_k}\;
\frac{1}{2 I_k}
\left\|\,\tilde{U}_k\,S_k\,(I - W)
-\sum_{i=1}^p M_i\,\tilde{U}_k\,S_k\,A^{(p)} 
\right\|_F^2
\;+\;
\frac{\rho_k}{2}\;\Bigl\|\,
U_k^{(t+1)} \;-\;\tilde{U}_k + \mu_{\tilde{U}_k}^{(t)}
\Bigr\|_F^2.
\]
To obtain the vectorized version, we have the following for the first term. 

\[
\tilde{U}_k\,S_k\,(I - W)
\;-\;
\sum_{i=1}^p
M_i\,\tilde{U}_k\,S_k\,A^{(p)} 
\;\;=\;\;
\tilde{U}_k\,S_k\,(I-W)
\;-\;
\sum_{i=1}^p
\bigl(M_i\,\tilde{U}_k\,S_k\,A^{(p)}\bigr).
\]
For \(\tilde{U}_k\,S_k\,(I - W)\):  
   \[
   \Bigl[\,(I-W)^\top S_k^\top \otimes I \Bigr]\mathbf{u}_k
   \]
where \(\mathbf{u}_k = \text{vec}(\tilde{U}_k)\). For \(M_i\,\tilde{U}_k\,S_k\,A^{(p)}\):
   \[
     \text{vec}(M_i\,\tilde{U}_k\,S_k\,A^{(p)})
     \;=\;
     \bigl[{A^{(p)}}^\top \otimes M_i\bigr]
     \,\text{vec}\bigl(\tilde{U}_k\,S_k\bigr)
     \;=\;
     \bigl[{A^{(p)}}^\top S_k^\top \otimes M_i\bigr]
     \mathbf{u}_k.
   \]
Hence the entire difference inside the norm becomes (in vector form):
\[
\Bigl[\,(I-W)^\top S^\top \otimes I \Bigr]\mathbf{u}_k
\;-\;
\sum_{i=1}^p
\bigl[{A^{(p)}}^\top S_k^\top \otimes M_i\bigr]\mathbf{u}_k.
\]
We can define 

\[
\Phi = \,(I-W)^\top S^\top \otimes I- \sum_{i=1}^p {A^{(p)}}^\top S_k^\top \otimes M_i
\]
Hence, the first term is 

\[
\frac{1}{2\,I_k}\;\bigl\|\Phi\,\mathbf{u}_k\bigr\|_2^2.
\]
The second penalty term is

\[
\frac{\rho_k}{2}\;\Bigl\|\,
U_k^{(t+1)} \;-\;\tilde{U}_k + \mu_{\tilde{U}_k}^{(t)}
\Bigr\|_F^2.
\]
Vectorizing:
\[
\text{vec}\bigl(U_k^{(t+1)} - \tilde{U}_k + \mu_{\tilde{U}_k}^{(t)}\bigr)
\;=\;
\text{vec}(U_k^{(t+1)}) 
\;-\; \mathbf{u}_k 
\;+\; \text{vec}\bigl(\mu_{\tilde{U}_k}^{(t)}\bigr).
\]
Define
\[
\mathbf{v}_k^{(t)} 
\;=\;
\text{vec}\bigl(U_k^{(t+1)} + \mu_{\tilde{U}_k}^{(t)}\bigr),
\]
so that term becomes
\[
\frac{\rho_k}{2}\;\Bigl\|\,
\mathbf{v}_k^{(t)} \;-\; \mathbf{u}_k
\Bigr\|_2^2.
\]
Putting both parts together gives:

\[
\underbrace{\frac{1}{2\,I_k}\;\bigl\|\Phi\,\mathbf{u}_k\bigr\|_2^2}_{\text{first term}}
\;+\;
\underbrace{\frac{\rho_k}{2}\,\bigl\|\mathbf{v}_k^{(t)} - \mathbf{u}_k\bigr\|_2^2}_{\text{second term}}.
\]
Rewriting \(\|\Phi\,\mathbf{u}_k\|_2^2 = \mathbf{u}_k^\top\,(\Phi^\top \Phi)\,\mathbf{u}_k\), the objective is

\[
\frac{1}{2\,I_k}\;\mathbf{u}_k^\top \Phi^\top \Phi\,\mathbf{u}_k
\;+\;
\frac{\rho_k}{2}\;\|\mathbf{v}_k^{(t)} - \mathbf{u}_k\|_2^2.
\]
Since we want to minimize
\(\displaystyle
\frac{1}{2\,I_k}\;\mathbf{u}_k^\top(\Phi^\top \Phi)\,\mathbf{u}_k
\;+\;
\frac{\rho_k}{2}\;\|\mathbf{v}_k^{(t)} - \mathbf{u}_k\|_2^2.
\)
Take derivative w.r.t. \(\mathbf{u}_k\) and set it equal to zero:

\[
\mathbf{u}_k
\;=\;
\Bigl(\frac{1}{I_k}\,\Phi^\top \Phi \;+\;\rho_k\,I\Bigr)^{-1}\,\rho_k\,\mathbf{v}_k^{(t)}.
\]
We can reshape the vector back to matrix as 
\[
\tilde{U}_k
\;=\;
mat\bigl[\Bigl(\frac{1}{I_k}\,\Phi^\top \Phi \;+\;\rho_k\,I\Bigr)^{-1}\,\rho_k\,\mathbf{v}_k^{(t)}\bigr].
\]

\subsection{Closed form of H}
\label{Up_H}

The gradient of the Frobenius norm term \(\| A - Q_k H \|_F^2\) with respect to \( H \) is:
   \[
   \nabla_H = \rho_k Q_k^\top (Q_k H - (U_k^{(t+1)} + \mu_{\hat{U}_k}^{(t)})).
   \]
   Summing over all \( k \) and setting the gradient to zero:
   \[
   \sum_{k=1}^K \rho_k Q_k^\top Q_k H = \sum_{k=1}^K \rho_k Q_k^\top (U_k^{(t+1)} + \mu_{\hat{U}_k}^{(t)}).
   \]

Apply Orthogonality Constraint (\( Q_k^\top Q_k = I \)):  
   Substitute \( Q_k^\top Q_k = I \):
   \[
   \left( \sum_{k=1}^K \rho_k \right) H = \sum_{k=1}^K \rho_k Q_k^\top (U_k^{(t+1)} + \mu_{\hat{U}_k}^{(t)}).
   \]

\subsection{Closed form for $S_k$}
\label{Up_S}
For computing the closed-Form, we have: 
\[
\min_{\mathbf{s}_k} \left\| \mathbf{x}_k - (V \odot U_k) \mathbf{s}_k \right\|_2^2 + \frac{\rho_d}{2} \left\| \mathbf{s}_k - (\tilde{\mathbf{s}}_k - \boldsymbol{\mu}) \right\|_2^2.
\]  
Setting the gradient with respect to \(\mathbf{s}_k\) to zero:  
\[
\left( (V \odot U_k)^\top (V \odot U_k) + \frac{\rho_d}{2} I \right) \mathbf{s}_k = (V \odot U_k)^\top \mathbf{x}_k + \frac{\rho_d}{2} (\tilde{\mathbf{s}}_k -\boldsymbol{\mu}).
\]  
By using the identity \((V \odot U_k)^\top (V \odot U_k) = VV^\top * U_k^\top U_k\), the solution becomes:  
\[
\mathbf{s}_k = \left( V^\top V * U_k^\top U_k + \frac{\rho_d}{2} I \right)^{-1} \left( \text{vec}(U_k^\top X_k V) + \frac{\rho_d}{2} (\tilde{\mathbf{s}}_k - \boldsymbol{\mu}) \right).
\]  

\subsection{Closed form of $\tilde{S}_k$}
\label{Up_tS}
We have the problem as below:

\[
\min_{\tilde{S}_k}
\;\;
f_{S_k}(\tilde{S}_k)
\;+\;
\frac{\rho_k}{2}
\|\tilde{S}_k \;-\;Q_k\|_{F}^{2},
\]
where
\[
f_{S_k}(\tilde{S}_k)
\;=\;
\frac{1}{2\,I_k}\,\Bigl\|\,
U_k\,\tilde{S}_k
\;-\;
U_k\,\tilde{S}_k\,W
\;-\;
\sum_{i=1}^{p} U_k^{\,I_k - i}\,\tilde{S}_k\,A^{(p)}
\Bigr\|_{F}^{2},
\quad
Q_k \;=\; S_k^{(t+1)} \;+\;\mu_{\tilde{S}_k}^{(t)}.
\]
We introduce the vectorized variables

\[
\mathbf{s}_k 
\;=\;
\mathrm{vec}\bigl(\tilde{S}_k\bigr),
\quad
\mathbf{q}_k
\;=\;
\mathrm{vec}\bigl(Q_k\bigr).
\]
We will vectorize this prolem for solving the closed form: 
\begin{itemize}
    \item \(\mathrm{vec}\bigl(U_k\,\tilde{S}_k\bigr)\)  
   Using \(A = U_k,\,X = \tilde{S}_k,\,B = I\), we get
   \[
     \mathrm{vec}\bigl(U_k\,\tilde{S}_k\bigr) 
     \;=\; \bigl(I^{T}\,\odot\,U_k\bigr)\,\mathbf{s}_k 
     \;=\; \bigl(I\,\odot\,U_k\bigr)\,\mathbf{s}_k
     \quad(\text{since }I^\top=I).
   \]
   \item \(\mathrm{vec}\bigl(U_k\,\tilde{S}_k\,W\bigr)\)  
   Here \(A = U_k,\,X = \tilde{S}_k,\,B = W\).  Thus
   \[
     \mathrm{vec}\bigl(U_k\,\tilde{S}_k\,W\bigr)
     \;=\; \bigl(W^{T}\,\odot\,U_k\bigr)\,\mathbf{s}_k.
   \]
   \item \(\mathrm{vec}\bigl(U_k^{I_k - i}\,\tilde{S}_k\,A^{(p)}\bigr)\) for each \(i\).  
   We have \(A = U_k^{I_k - i},\,X = \tilde{S}_k,\,B = A^{(p)}\).  So
   \[
     \mathrm{vec}\bigl(U_k^{I_k - i}\,\tilde{S}_k\,A^{(p)}\bigr)
     \;=\; \bigl({A^{(p)}}^{T}\,\odot\,U_k^{I_k - i}\bigr)\,\mathbf{s}_k.
   \]
\end{itemize}

Hence the entire quantity inside the Frobenius norm  
\(\;U_k\,\tilde{S}_k \;-\; U_k\,\tilde{S}_k\,W \;-\; \sum_{i=1}^{p} U_k^{I_k - i}\,\tilde{S}_k\,A^{(p)}\;\)  
becomes a linear operator in \(\mathbf{s}_k\).  Concretely,

\[
\mathrm{vec}\Bigl(
U_k\,\tilde{S}_k
\;-\;
U_k\,\tilde{S}_k\,W
\;-\;
\sum_{i=1}^{p} U_k^{I_k - i}\,\tilde{S}_k\,A^{(p)}
\Bigr)
\;=\;
\underbrace{
\bigl(I\odot U_k\bigr)
}_{\text{Term 1}}
\mathbf{s}_k
\;-\;
\underbrace{
\bigl(W^{T}\!\odot\,U_k\bigr)
}_{\text{Term 2}}
\mathbf{s}_k
\;-\;
\sum_{i=1}^{p}
\underbrace{
\bigl({A^{(p)}}^{T}\odot\,U_k^{I_k - i}\bigr)
}_{\text{Term 3}}
\mathbf{s}_k = T_k\,\mathbf{s}_k
\]
where,

\[
T_k
\;=\;
\bigl(I\odot U_k\bigr)
\;-\;
\bigl(W^{T}\odot U_k\bigr)
\;-\;
\sum_{i=1}^{p}
\bigl({A^{(p)}}^{T}\odot U_k^{I_k - i}\bigr).
\]

Thus the original objective becomes

\[
\frac{1}{2\,I_k}\,\|\,T_k\,\mathbf{s}_k\,\|_{2}^{2}
\;+\;
\frac{\rho_k}{2}\,\|\mathbf{s}_k \;-\;\mathbf{q}_k\|_{2}^{2},
\quad\text{where}\quad
\mathbf{s}_k = \mathrm{vec}(\tilde{S}_k),
\;\;
\mathbf{q}_k = \mathrm{vec}(Q_k).
\]
By taking the derivative, we can have: 

\[
\nabla_{\mathbf{s}_k}\,g(\mathbf{s}_k)
\;=\;
\frac{1}{I_k}\,T_k^{T}\,T_k\,\mathbf{s}_k
\;+\;
\rho_k\,\bigl(\mathbf{s}_k - \mathbf{q}_k\bigr).
\]
Setting this to zero yields

\[
\frac{1}{I_k}\,T_k^{T}\,T_k\,\mathbf{s}_k
\;+\;
\rho_k\,\mathbf{s}_k
\;=\;
\rho_k\,\mathbf{q}_k.
\]
so

\[
\boxed{
\mathbf{s}_k
\;=\;
\Bigl(\tfrac{1}{I_k}\,T_k^{T}\,T_k + \rho_k\,I\Bigr)^{-1}
\;\bigl(\rho_k\,\mathbf{q}_k\bigr).
}
\]
where 
\[
T_k
\;=\;
\bigl(I\odot U_k\bigr)
\;-\;
\bigl(W^{T}\odot U_k\bigr)
\;-\;
\sum_{i=1}^{p}
\bigl({A^{(p)}}^{T}\odot U_k^{I_k - i}\bigr).
\]
and 
\[
\mathbf{q_k}= \mathrm{vec}\bigl(S_k^{(t+1)} +\mu_{\tilde{S}_k}^{(t)}\bigr).
\]

\section{Simulation Data Generating:}
\label{sim}
\textbf{Intra-slice graph:} We use the \textit{Erdős-Rényi (ER) model} to generate a random, directed acyclic graph (DAG) with a target mean degree \(pr\). In the ER model, edges are generated independently using i.i.d. Bernoulli trials with a probability \(pr/dr\), where \(dr\) is the number of nodes. The resulting graph is first represented as an adjacency matrix and then oriented to ensure acyclicity by imposing a lower triangular structure, producing a valid DAG. Finally, the nodes of the DAG are randomly permuted to remove any trivial ordering, resulting in a randomized and realistic structure suitable for downstream applications.\\
\textbf{Inter-slice graph:} We still use \textit{ER model} to generate the weighted matrix. The edges are directed from node \( i_{t-1} \) at time \( t-1 \) to node \( j_t \) at time \( t \). The binary adjacency matrix \( A_{\text{bin}} \) is constructed as:
\[
A_{i_{t-1}, j_{t}} = 
\begin{cases}
1 & \text{with probability } pr/dr \quad \text{for edges from node } i_{t-1} \text{ to } j_t, \\
0 & \text{otherwise}.
\end{cases}
\]
\textbf{Assigning Weights}: 
Once the binary adjacency matrix is generated, we assign edge weights from a \textit{uniform distribution} over the range \([-0.5, -0.3] \cup [0.3, 0.5]\) for \(W\) and \([-0.5\alpha, -0.3\alpha] \cup [0.3\alpha, 0.5\alpha]\) for \(A\), where:
\[
\alpha = \frac{1}{\eta^{p-1}},
\]
and \( \eta \geq 1 \) is a decay parameter controlling how the influence of edges decreases as time steps get further apart.

\section{Causal Phenotype Network procedure}
\label{CPN proc}
\begin{figure}[H]
  \centering
  \begin{subfigure}[b]{0.45\textwidth}
    \centering
    \includegraphics[width=\textwidth]{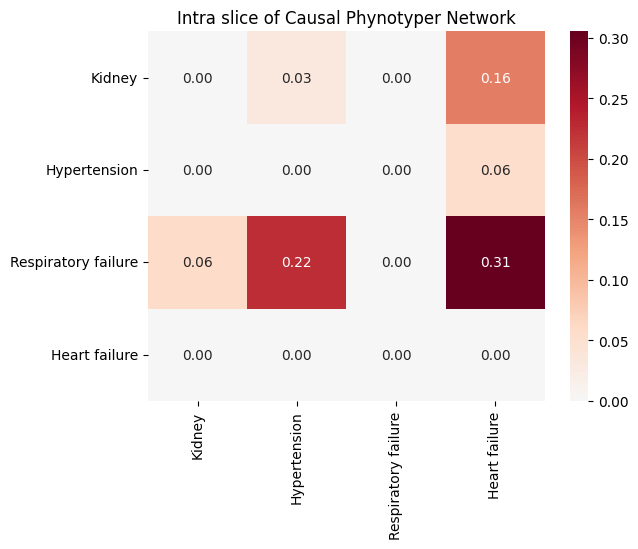}
    \caption{}
  \end{subfigure}
  \begin{subfigure}[b]{0.45\textwidth}
    \centering
    \includegraphics[width=\textwidth]{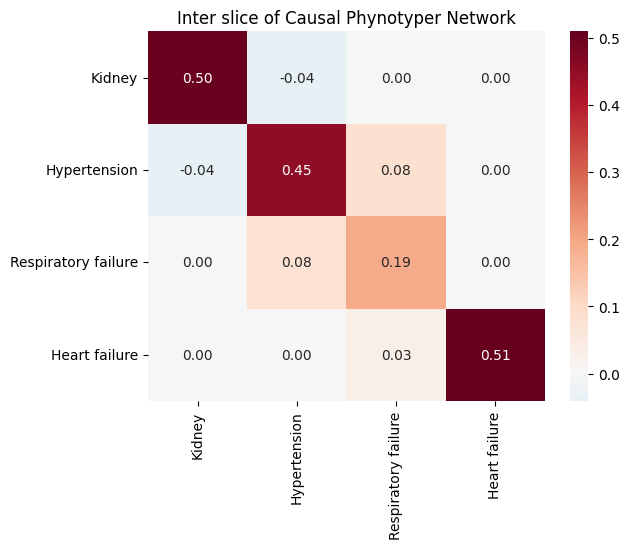}
    \caption{}
  \end{subfigure}
 \caption{An example for causal phenotype network generated by \texttt{CaRTeD}}
\label{fig_app:cpn_eg}
\end{figure}

As shown in our paper, we will have two causal graphs in heatmap form, one for intra-slice \(W\) and inter-slice \(A\). The heatmaps generated by \texttt{CaRTeD} have been shown in Fig.~\ref{fig_app:cpn_eg}. Since our causal-discovery method does not perform explicit causal-effect inference, we convert the resulting directed graph into a causal diagram. In our framework, \(A\) serves as the complementary information matrix; for example, it supplies edges that are absent in \(W\). Therefore, we observe two additional causal edges (i.e., two off-diagonal entries) contributed by \(A\).

\end{document}